\documentclass[10pt,twocolumn,letterpaper]{article}

\usepackage{CVPR2026/cvpr}              
\definecolor{cvprblue}{rgb}{0.21,0.49,0.74}
\usepackage[pagebackref,breaklinks,colorlinks,allcolors=cvprblue]{hyperref}

\def\paperID{44096} 
\def\confName{CVPR}
\def\confYear{2026}

\title{Make it SING: Analyzing Semantic Invariants in Classifiers}

\author{Harel Yadid, Meir Yossef Levi, Roy Betser, Guy Gilboa\\
Viterbi Faculty of Electrical and Computer Engineering\\
Technion -- Israel Institute of Technology, Haifa, Israel\\[4pt]
{\tt\small \{harel.yadid,roybe,me.levi\}@campus.technion.ac.il; guy.gilboa@ee.technion.ac.il}
}

\usepackage{cuted}
\usepackage[T1]{fontenc}
\usepackage{courier}
\usepackage{titletoc}   
\usepackage{listings}
\addtocontents{toc}{\protect\setcounter{tocdepth}{-10}}

\begin{document}

\maketitle
\begin{strip}
\centering  \includegraphics[width=0.90\textwidth]{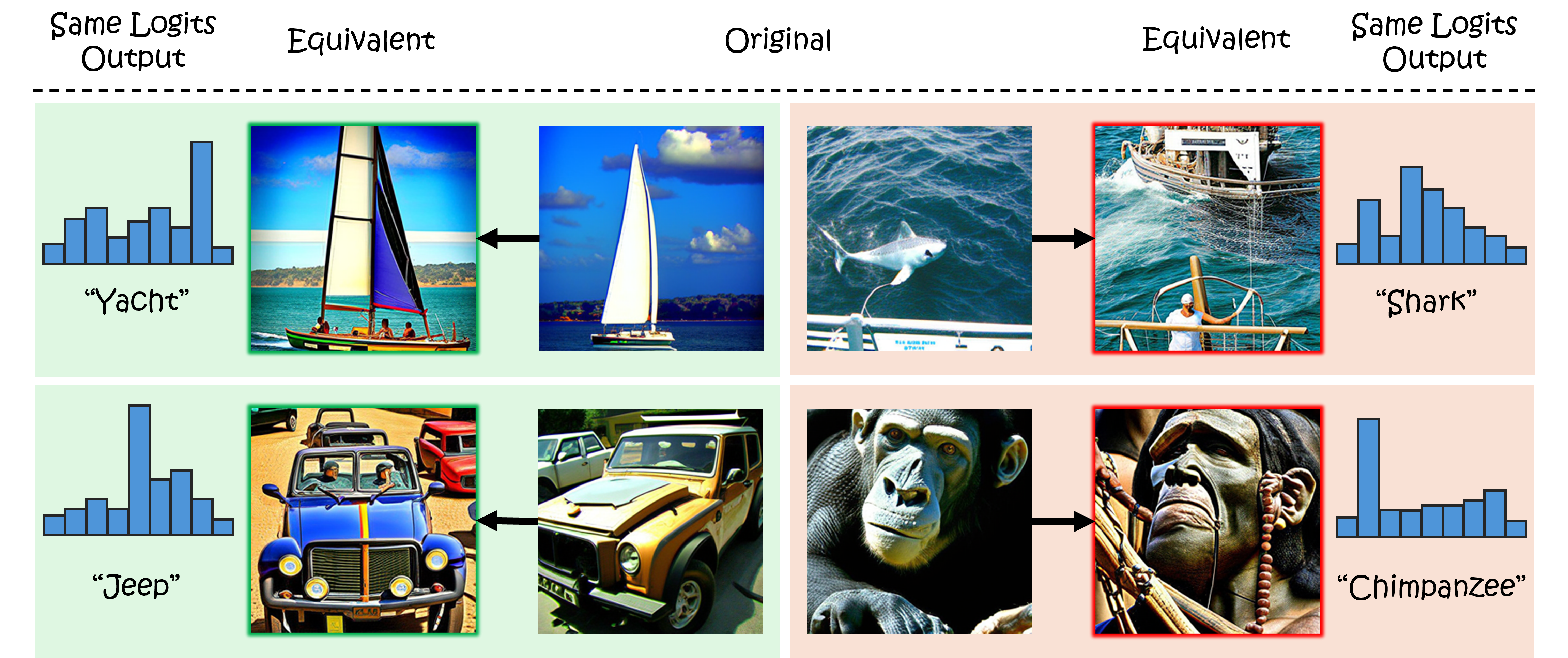}
\captionsetup{hypcap=false}
\captionof{figure}{{\bf Visualization of benign and problematic invariants.} The four images at the center correspond to certain features taken from a pretrained ResNet50. On the left and right columns their equivalent images are shown, following null-space removal. Each pair yields the same logits after passing through the linear head. The left side (green) demonstrates robustness, with little semantic change. The right side (red) incurs large semantic deviations. Our framework quantifies these changes statistically, diagnosing semantic invariants at the class and network level.} 
  \label{fig:subspace_figure}
\end{strip}


\begin{abstract}
All classifiers, including state-of-the-art vision models, possess invariants, partially rooted in the geometry of their linear mappings. These invariants, which reside in the null-space of the classifier, induce equivalent sets of inputs that map to identical outputs. The semantic content of these invariants remains vague, as existing approaches struggle to provide human-interpretable information. To address this gap, we present Semantic Interpretation of the Null-space Geometry (SING), a method that constructs equivalent images, with respect to the network, and assigns semantic interpretations to the available variations. We use a mapping from network features to multi-modal vision language models. This allows us to obtain natural language descriptions and visual examples of the induced semantic shifts. SING can be applied to a single image, uncovering local invariants, or to sets of images, allowing a breadth of statistical analysis at the class and model levels. For example, our method reveals that ResNet50 leaks relevant semantic attributes to the null space, whereas DinoViT, a ViT pretrained with self-supervised DINO, is superior in maintaining class semantics across the invariant space. \thanks{Code is available at \url{https://tinyurl.com/github-SING}.}

\end{abstract}

\section{Introduction}
\label{sec:intro}
State of the art networks, especially vision classifiers, learn internal representations with complex geometry. while this correlates with strong performance on recognition benchmarks, it makes mechanistic interpretability difficult \citep{doshivelez2017rigorous,ansuini2019intrinsicdimensiondatarepresentations}.
For example, invariants, derived from the null space of the model's linear layers, lead to sets of inputs with identical outputs. We refer to these sets as \emph{equivalent sets}. Whereas nonsemantic invariants such as background or illumination are generally beneficial, invariants that carry semantic information may harm the classifier. However, although users can often introduce image augmentations to increase invariants of certain attributes, they cannot easily determine what the model has actually learned, only via rigorous testing. 

This motivates approaches that interpret neural networks while focusing on their geometry. A natural starting point would be the geometry of the classification head, where the last decision is made. A related line of research applies singular value decomposition (SVD) to the latent space based on representative data in the latent feature space \citep{aubry2015understanding,Harkonen2020,haas2024discovering}; however, these methods are prone to the data covariances rather than network mechanism.
Other methods operate directly in the weight-induced null space \citep{cook2020outlier,Rezaei2023_NullSpaceExplain,li2024nullspace}. For example, the classifier head can be decomposed into two space components:(i) principal directions, associated with dominant singular values that influence the logits; (ii) null directions, the complementary space that keeps the inputs unchanged \citep{Praggastis2022_SVDWeights,anthes2023keep}. While they are able to identify the existence of invariant directions, they fail to explain semantically what they represent, and often rely on task-specific data to demonstrate these directions \citep{li2024nullspace}.

Recent advances in mechanistic interpretability \citep{moayeri2023text2concept,kim2023grounding,huang2024lg,dreyer2025mechanistic} leverage the translation of latent features from a given model into a multi-modal vision language space, most notably CLIP \citep{radford2021clip}. The use of CLIP to compute semantic correlations between text and images facilitates new sets of techniques that focus on producing human-readable concepts and counterfactual examples to aid interpretation. However, to the best of our knowledge, we are the first to map a classifier’s invariant directions into a multi modal network for systematic analysis, providing textual descriptions and visual examples.

We propose a Semantic Interpretation of the Null-space Geometry (SING), a method grounded in SVD of the feature layer to probe the latent feature space of a target classifier and identify the representations of equivalent pairs. The revealed null-space structure is then mapped to CLIP's vision-language space through linear translators, yielding quantifiable semantic analysis.
Our method provides a general framework for measuring human-readable explanations of data invariants, spanning from image and class levels up to entire model assessments. It supports probing, debugging, and comparing these invariants across vulnerable classes and spurious correlations such as background cues, as well as measuring how much a specific concept is ignored by the model.
We demonstrate the effectiveness of SING through cross-architecture measurements, per-class analysis, and individual image breakdown. In the last section of our experiments we present a promising direction for null space manipulation, creating features with hidden semantics that the model ignores. 
Our main contributions are:
\begin{itemize}
    \item \textit{A semantic tool for interpreting invariants}. SING links classifier geometry, specifically the null space and the invariants it induces, to meaningful human-readable explanations using equivalent pairs analysis.
    
    \item \textit{Model comparison}. We introduce a protocol to compare different architectures by measuring the leakage of their semantic information into their null space. 
    Our analysis found that DinoViT, among the examined networks, had the least class-relevant leakage into its null space while allowing broad permissible invariants, such as background or color.
    
    \item \textit{Open vocabulary class analysis}. Our framework allows for systematic investigations of the sensitivity of classes to certain concepts. It can discover spurious correlations and assess their contribution. For example, our experiments show that for some spurious attributes in the DinoViT model the classifier head considers them as invariants.  
\end{itemize}
\begin{figure*}[t]
  \centering
  \includegraphics[width=\textwidth]{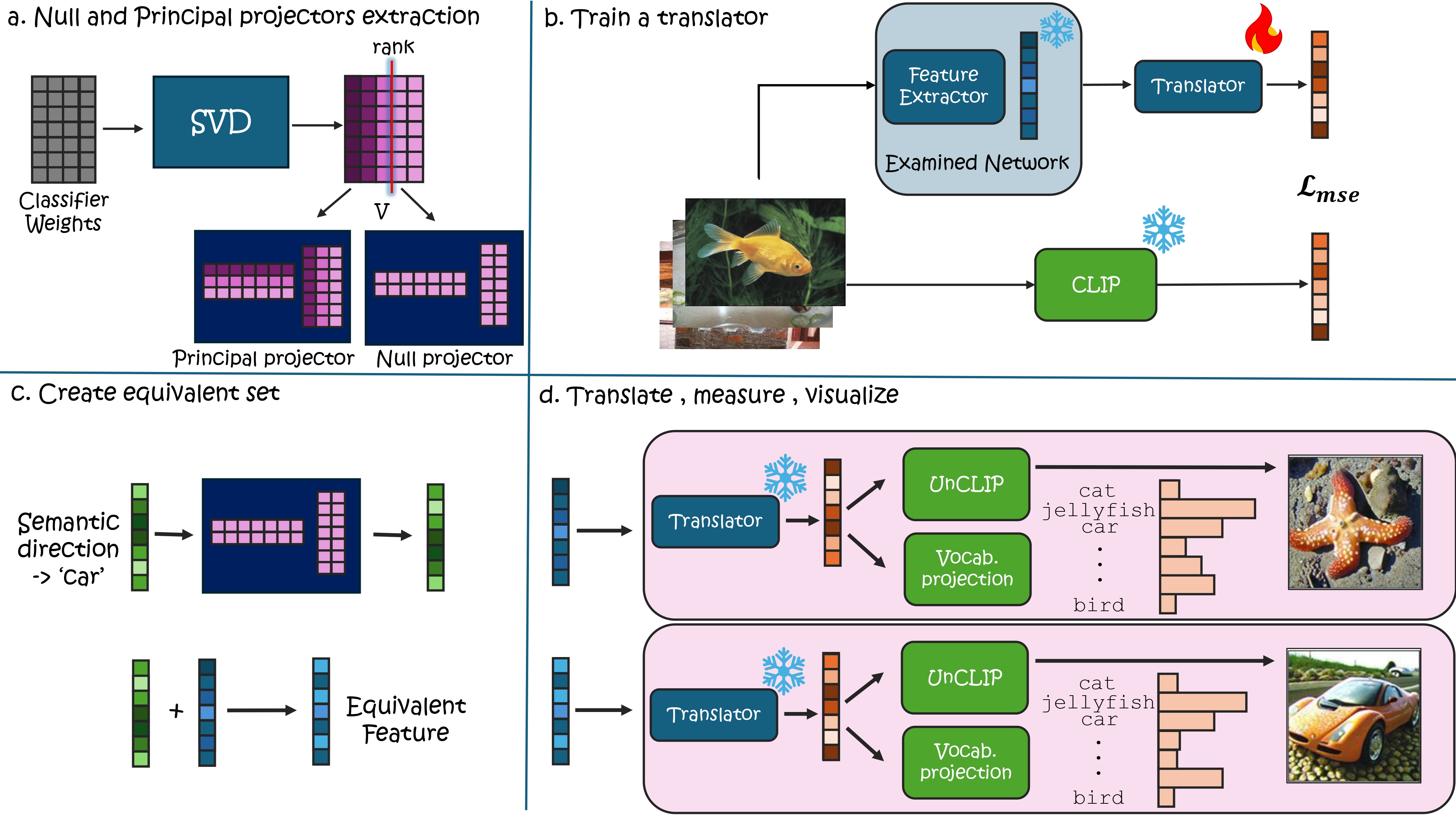}
  \caption{{\bf Method Overview.} The approach consists of: (a) decomposing the final linear weights to obtain principal and null projectors; (b) training a translator that maps features from the network embedding space to the CLIP image space; (c) creating an equivalent pair to the feature we want to examine. (d) translate the set into CLIP image embedding space, and apply our metrics and visualizations.
  }
  \label{fig:method}
\end{figure*}


\section{Related Work}
\subsection{Explainability through decomposition}

Decomposing latent spaces using SVD is a foundational approach for studying their invariances \citep{GolubReinsch1970}. \citet{aubry2015understanding} used this technique to probe dominant modes of variation in CNN embeddings, for example illumination and viewpoint, under controlled synthetically rendered scenes. \citet{Harkonen2020} applied it to GAN latent spaces for interpretable controls, and more recently \citet{haas2024discovering} used it to present consistent editing directions in diffusion model latent spaces.
However, feature-space decomposition is inherently data-dependent: its axes reflect the covariance of the measured dataset rather than the classifier’s decision geometry. Notably, it may miss invariants residing in the classifier’s null space itself. 

A complementary study involves decomposing the model weights directly. This line of work includes early low-rank decompositions of convolutional weights for acceleration \citep{Jaderberg2014_ConvCompression}, SVD analyzes of convolutional filters for interpretability \citep{Praggastis2022_SVDWeights}, and decomposition of the final linear layer to identify the direction relevant to the task and the direction invariant to the task \citep{anthes2023keep}.
Null space analysis has been explored across several directions in deep learning. Some works leverage it for information removal: \citet{ravfogel2020null} iteratively projected representations onto the null space of a linear attribute classifier to remove protected information while preserving task predictions, while \citet{li2024nullspace} exploited null space properties for image steganography, masking images that leave logits unchanged. Others use it as a diagnostic tool: \citet{cook2020outlier} derived OOD detection scores from null space projections, and \citet{idnani2023don} explained OOD failures via null-space occupancy, showing that features drifting into the readout's null space lead to misclassification. \citet{Rezaei2023_NullSpaceExplain} further analyzed the last layer null space to quantify overfitting through changes in its structure.
Collectively, these methods treat the null space as an operational invariance set for control, detection, and manipulation. However, as far as we know, no current research managed to assign \emph{semantic meaning} to null directions, as our approach does.

\subsection{Projecting features to a vision-language space}
Contrastive Language–Image Pretraining (CLIP) \citep{radford2021clip} learns a rich joint embedding space for images and text, enabling a wide range of vision-language applications. 
A characteristic property of this space is the presence of a modality gap between image and text embeddings~\citep{liang2022mind}. 
Beyond its empirical success, the geometry of the CLIP latent space has been studied from multiple perspectives, including geometric analyses~\citep{levi2025double}, probabilistic modeling~\citep{betser2025whitenedclip,betser2026general}, and asymptotic theoretical analysis~\citep{betserinfonce}.
Several methods have leveraged CLIP representations for interpretability, either by mapping classifier features into CLIP's vision-language space or by using CLIP as supervision to train concept vectors within the target model's feature space.
Text2Concept \citep{moayeri2023text2concept} learns a linear map from any vision encoder to CLIP's space, turning text prompts directly into concept activation vectors, while CounTEX \citep{kim2023grounding} introduces a bidirectional projection between classifier and CLIP to generate counterfactual explanations. CLIP-Dissect \citep{oikarinen2022clip} extends this direction to the neuron level, automatically assigning open-vocabulary concept labels to individual neurons by matching their activation patterns to CLIP embeddings.
Rather than projecting into CLIP, LG-CAV \citep{huang2024lg} uses CLIP's text-image scores on unlabeled probe images as supervision to train concept vectors directly within the target model's feature space.
Taking a broader view, DrML \citep{zhang2023diagnosing}, MULTIMON \citep{tong2023mass}, and MDC \citep{chen2025model} use language to probe, mine, and correct vision model failures across a range of failure modes.
Despite the breadth of these approaches, they all focus on the active feature subspace of the classifier, leaving the null space unexplored.

\section{Method}
\label{sec:method}
Our method contains several components as can be seen in \Cref{fig:method}. We begin by decomposing the target layer into principal and null subspaces and building projection operators that isolate each space. On the second component, we learn a linear mapping that translates the layer’s features into the shared multi-modal space, specifically the image space. We then select a feature and perturb it along a specified semantic direction projected to a chosen subspace, creating the equivalent feature pair. After perturbing, we translate the feature using our translator to observe how its representation changed semantically with visualization and textual measurements. In this section we develop each component in detail, with particular attention to the null space and to the classifier head.
\subsection{Setup}
 In our work, we focus on the last fully connected layer $W \in \mathbb{R}^{c \times m}$, which maps the penultimate features $f \in \mathbb{R}^m$ to a logit vector in the dimension of the number of classes $c$. We decompose it with SVD and specifically extract the null space projection matrix $\Pi_\text{n}$, which contains all the invariants of the layer. In the translation step we denote $T_{\Theta}(f)$ as the \emph{Translator}, and we use CLIP as our multi-modal model space. We denote $z^{img}$ and $z^{text}$ as the image and text latent features in CLIP space. We define $\tilde{f}$ as the equivalent pair of $f$ after perturbation in the null space.
 
\subsection{SVD on the classifier head}
$W$ can be decomposed into its principal and null spaces via SVD:

\begin{equation}
    W = U \, \Sigma \, V^\top, \qquad
V = \bigl[\,V_{\mathrm{p}} \ \ V_{\mathrm{n}}\,\bigr],
\end{equation}

where $\Sigma \in \mathbb{R}^{c \times m}$ is a rectangle diagonal matrix containing the singular values in descending order, 
and $U \in \mathbb{R}^{c \times c}$ and $V \in \mathbb{R}^{m \times m}$ contain the left and right singular vectors, respectively.
We take $\operatorname{rank}(W)$, and use it to break the right singular vectors $V$ into the two subspace components, \emph{principal space}, denoted $V_{\mathrm{p}}$ (associated with non-zero singular values),  and the remaining columns $V_{\mathrm{n}}$ that span the \emph{null space}.
Any perturbation $\nu \in \operatorname{span}(V_{\mathrm{n}})$ leaves the logits unchanged:
\begin{equation}
W(f + \nu) = Wf + W\nu = Wf,
\end{equation}
since $W\nu = 0$ for all $\nu$ in the null space.
Consequently, our projector matrices are:
\begin{equation}
\label{eq:proj}
    \Pi_\text{p} = V_{\mathrm{p}} V_{\mathrm{p}}^{\top}, \qquad
    \Pi_\text{n} = V_{\mathrm{n}} V_{\mathrm{n}}^{\top}.
\end{equation}

\subsection{Training a translator}
Following \citet{moayeri2023text2concept} and justified by \citet{lahner2024direct}, we define a linear mapping operator $T:\mathbb{R}^{m}\to\mathbb{R}^{n}$. Recall that $f\in\mathbb{R}^{m}$ is the classifier feature and $z^{img}\in\mathbb{R}^{n}$ the corresponding image feature in CLIP. We fit $T_\Theta$ for a certain pretrained model by minimizing a loss combining mean squared error, and weight decay:

\begin{equation}
\mathcal{L} =\|T_{\Theta}(f) - z^{img}\|_2^2
+  \lambda \,\|\Theta\|_2^2,
\label{eq:translator_loss}
\end{equation}

where $\Theta$ is the parameters of the translator and $\lambda$ is a balancing coefficient. Detailed explanations on the training procedure can be found in the supplementary materials.
Note that since the translator is linear, it admits
$T_{\Theta}(f + v) = T_{\Theta}(f) + T_{\Theta}(v)$ for any $f,v$, hence naturally fits additive feature decompositions, as our framework suggests. The translator is validated to preserve relative classification performance across models, and while we use CLIP as the target space, we demonstrate in the supplementary that other vision-language models can serve this role as well. Although our framework is not limited to linear translators, we empirically verified that this linear map fits well in our setting.


\subsection{Metrics}
\paragraph{Attribute score.}
An angle between two nonzero vectors $x,y$ of the same dimension is defined by:
\begin{equation}
    \angle(x,y) := \arccos\bigr(\frac{x\cdot y}{\|x\|\|y\|}\bigl).
\end{equation}
CLIP Score, as described in \citet{hessel2021clipscore}, is the cosine similarity of the angle between a CLIP feature in image space $z^{img}$, and a feature in the text space, $z^{text}$. We write this angle as follows:
\begin{equation}
\angle(z^{img},z^{text})
\end{equation}

Recall that $f$ and $\tilde{f}$ are the original and its equivalent pair. We define \emph{Attribute Score} (AS) for text target $z^{text}$ as the difference between two angles:
\begin{equation}
\label{eq:AS}
    \text{AS}(f,\tilde{f}|z^{text},T_{\Theta}) := \angle(T_{\Theta}(f),z^{text}) - \angle(T_{\Theta}(\tilde{f}),z^{text}).
\end{equation}

A positive AS indicates that the equivalent image is semantically closer to the text and vice versa. In our framework, the text prompts are chosen as ``\texttt{an image of a \textless class\textgreater}''
 to analyze how null removal affects classification. However, this metric is general and can be applied with any prompt selection.
\paragraph{Image score.}  
While AS quantifies how the image deviates from its current semantics, the image may be altered in appearance without affecting AS. Such differences in overall appearance can be measured directly by the angular distance related to the original and its equivalent pair. we define it as \emph{Image Score} (IS):
\begin{equation}
\label{eq:IS}
    \text{IS}(f,\tilde{f}|T_{\Theta}) := \angle(T_{\Theta}(f), T_{\Theta}(\tilde{f})).
\end{equation}

Intuitively, AS captures the effects of null spaces on the alignment of text-image, whereas IS reflects general semantic changes in the image. When the text is in the correct image class we would like low AS, and hence null-space changes should not affect class distinction. However, a good classifier should allow high IS, and hence large semantic changes that do not affect class distinction, such as background change and other allowed semantic invariants.
Details on image synthesis for visualization are provided in the supplementary materials, however it's highly important to note that those visualizations are used only for qualitative illustration; all quantitative claims rely on logits and CLIP embeddings.
\subsection{Applications}
Our main focus is on removing the null component from an image feature $f$. This way, the equivalent pair is
\begin{equation}
\tilde{f} = f - \Pi_\text{n} f.
\label{eq:f_null}
\end{equation}
Both $f$ and $\tilde{f}$ produce the same logit vector under the examined network, yet the semantic content can be changed as a result of the null-removal process. In the following, we describe how to quantify semantic information leakage at different levels: model, attribute, and image, using the proposed metrics (AS and IS).
\paragraph{Model-level comparison.} A desirable property of well-performing classifiers is to maintain a rich invariant space, while ensuring that this richness does not compromise class preservation. For instance, there exists a wide variety of dogs differing in breed, pose, size, color, background and more, all of which should be classified consistently with high confidence. Hence, the invariant space should support such diversity. However, if perturbations along invariant directions lead to changes in classification confidence or even alter the predicted class, this indicates that class-specific information has leaked into the invariant space - a highly undesirable property that also exposes the model to adversarial vulnerabilities. To evaluate this, we collect a representative set of images (16 ImageNet classes, serving as a proof of concept), compute the AS and IS metrics (with respect to the real class prompt; ``\texttt{an image of a \textless ground-truth class\textgreater}'')  on all null-removed pairs, and perform a statistical analysis across models. An effective model should exhibit a broad range of IS values, reflecting rich invariance, while maintaining a narrow distribution of AS values, ensuring semantic consistency.

\paragraph{Class and Attribute analysis.} The same methodology can be applied to analyze inter-class behavior by selecting representative sets from different classes. We conducted two complementary variants. First, we collected images from each class independently and computed the absolute Attribute Score (AS) after null-removal, relative to the true label prompt. Higher AS values indicate that the classifier contains more semantic information within the invariant space for that class. This provides a practical diagnostic tool for practitioners when choosing networks suited to specific classes or domains. Second, we expanded the vocabulary to an open set of concepts. We quantified the distance (angles) between the original and the null-removed features, over a broad set of phrases,  revealing how semantic correlations emerge between the null space and diverse concepts.
\paragraph{Single image analysis.} Following the same logic, leakage can also be examined at the image level. This provides a fine-grained diagnostic tool for identifying and debugging failure cases.

\paragraph{Null perturbations.}
While null removal is useful for fair comparisons across classes, attributes, or images, feature manipulation need not be restricted to a single invariant direction. We propose a more principled selection of perturbation directions. We formalize perturbations that target a specific concept while remaining confined to the model’s invariant (null) subspace. 
Let $f\in\mathbb{R}^d$ be an image feature, $T_{\Theta}:\mathbb{R}^d\!\to\!\mathbb{R}^n$ the translator into the CLIP image-embedding space, and $z_{\text{text}}\in\mathbb{R}^n$ the CLIP text embedding of a prompt (e.g., ``\texttt{an image of a jellyfish}''). 
Define the cosine-similarity score
\begin{equation}
\label{eq:sim-score}
s(f; z_{\text{text}})\;:=\;\frac{ \langle z,\, z_{\text{text}}\rangle }{\|z\|\,\|z_{\text{text}}\|}\,, 
\qquad z := T_{\Theta}(f).
\end{equation}
The \emph{semantic direction} toward the prompt is the gradient through the translator,
\begin{equation}
\label{eq:grad-dir}
g_{\text{text}}(f)\;:=\;\nabla_{f}\, s(f; z_{\text{text}}).
\end{equation}
Let $\Pi_\text{n}$ denote the orthogonal projector onto the null space (\eqref{eq:proj}). Projecting this direction onto the null space isolates the component that lives in the invariant subspace:
\begin{equation}
\label{eq:null-proj}
d_{\text{null}}(f)\;:=\;P_{\mathcal{N}}\, g_{\text{text}}(f), 
\qquad 
\hat d_{\text{null}}(f)\;:=\;\frac{d_{\text{null}}(f)}{\|d_{\text{null}}(f)\|}.
\end{equation}
One can control the extent of semantic change via a scalar step size $\varepsilon$ applied to the normalized null direction $\hat d_{\text{null}}$:
\begin{equation}
\label{eq:null-stepj}
f_{\varepsilon}=f+\varepsilon\,\hat d_{\text{null}}(f).
\end{equation}

By choosing the prompt to correspond to another class or attribute, this construction probes a class’s sensitivity \emph{within} the invariant subspace to concepts associated with other classes, thereby revealing ``confusing'' inter-class relationships.

\section{Experiments}
\begin{figure}[tb]
  \centering
  \begin{subfigure}{0.48\textwidth}
  \includegraphics[width=\textwidth]{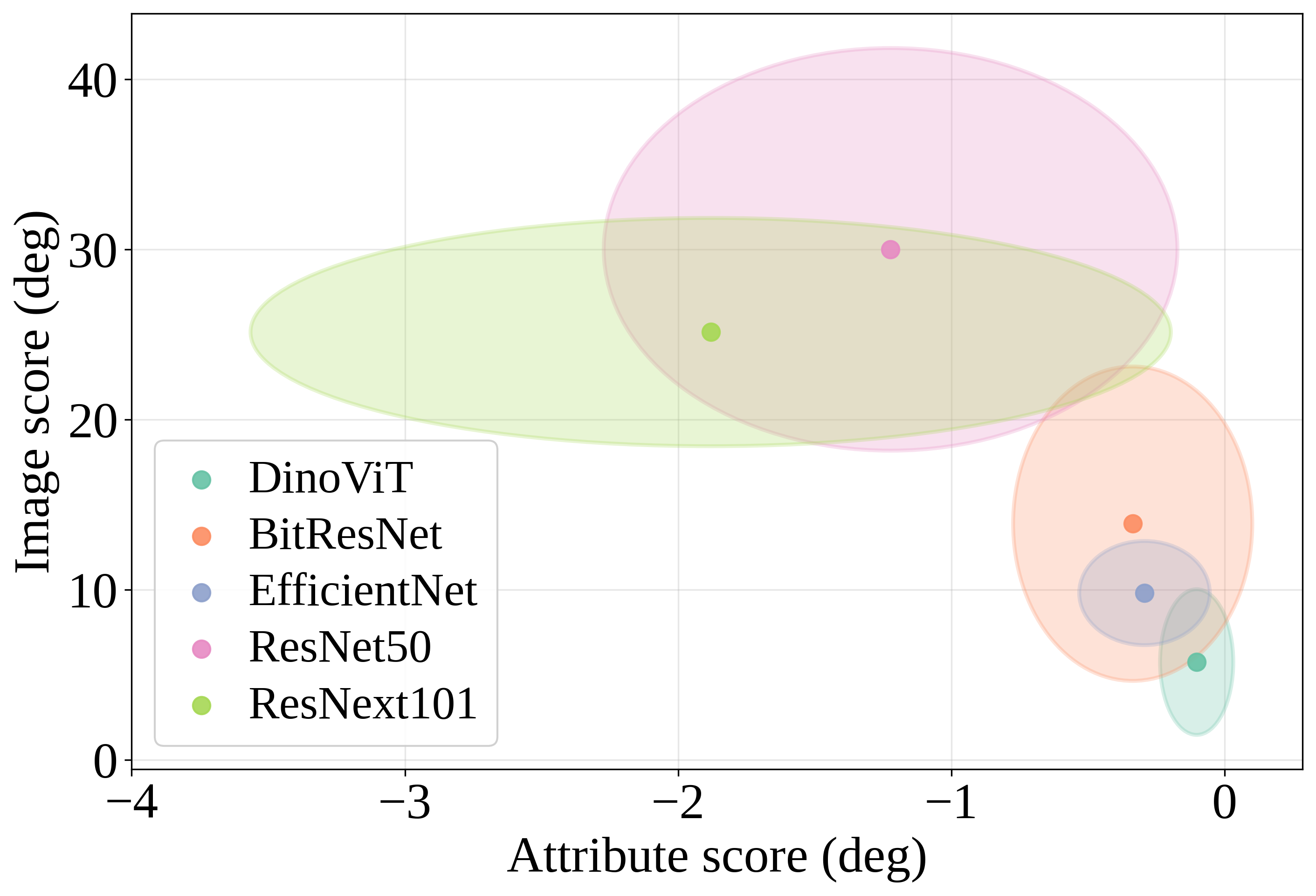}
  \caption{}
  \end{subfigure}
  \begin{subfigure}{0.48\textwidth}
  \includegraphics[width=\textwidth]{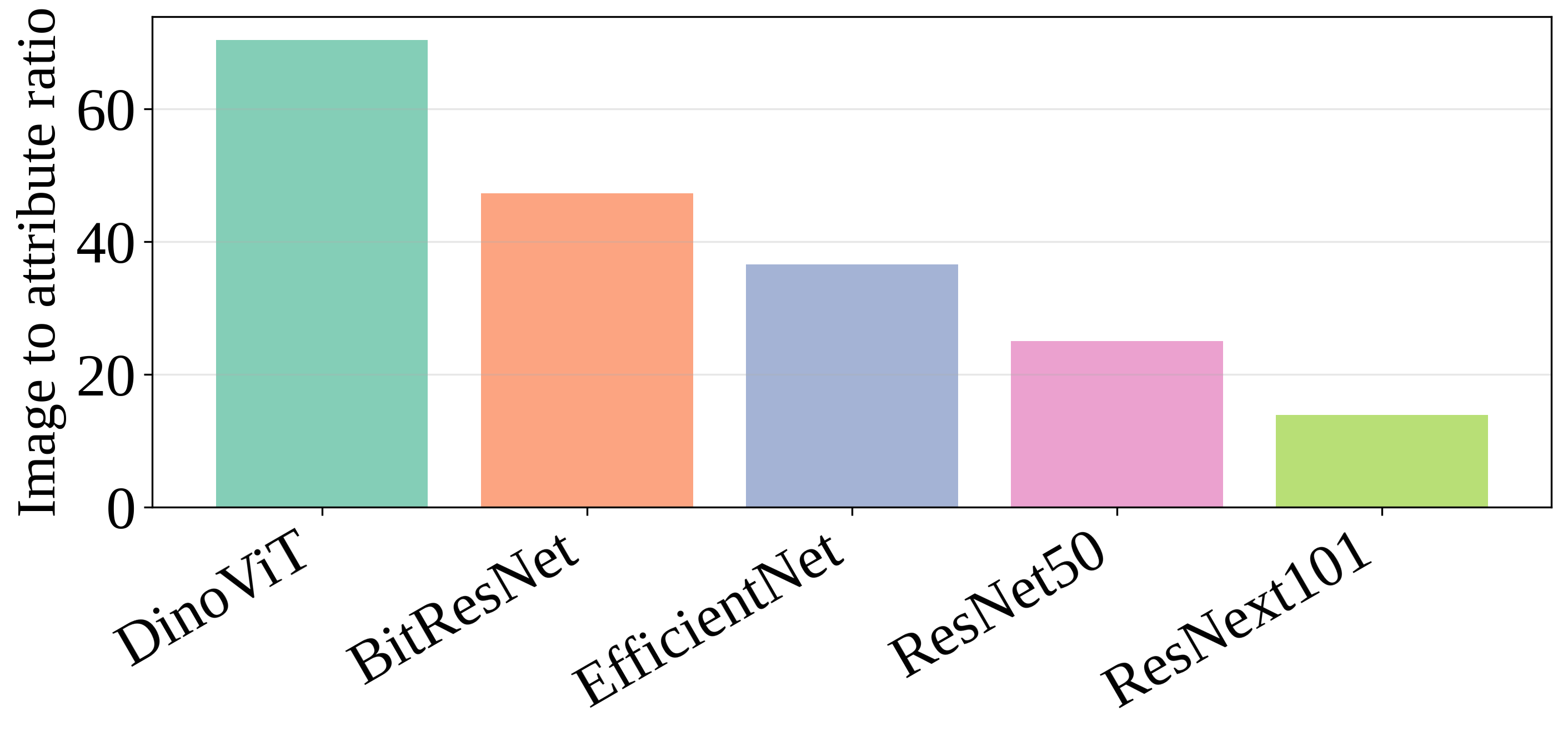}
  \caption{}
  \end{subfigure}
    
    \caption{{\bf Model-level comparison (1,000 classes).} (a) Attribute Score (AS) quantifies \emph{class-dependent} semantic leakage into the null space; Image Score (IS) quantifies tolerance to \emph{class-independent} (non--class-dependent) semantic variation within the invariant subspace. Desirably, AS is low and IS is high (relative to AS). In our results, DinoViT performs best in this regard. (b) We summarize the trade-off with the $\mathrm{IS}/\mathrm{AS}$ ratio (higher is better), DinoViT has the highest ratio and ResNext101 the lowest.}

    \label{fig:model_scatter}
\end{figure}

\subsection{Dataset and models}
We base our analysis on five models pretrained on ImageNet-1k \citep{deng2009imagenet} spanning diverse architectures and training paradigms: DinoViT \citep{caron2021emerging}, ResNet50 \citep{he2016deep}, ResNext101 with weakly supervised pretraining \citep{mahajan2018exploring}, EfficientNetB4 trained with Noisy Student \citep{xie2020self}, and BiTResNetv2 \citep{kolesnikov2020big}. For statistical analyses, we collect 10k feature vectors per model from all 1,000 ImageNet classes. For each model, we then train a dedicated translator in the same 1,000-class setting. We also empirically confirm that null-space removal leaves logits nearly unchanged, whereas equal-norm perturbations in other directions induce substantial logit and CLIP drift (see supplementary material).
\subsection{Model comparison}

We compare models globally across all tested classes, measuring AS and IS after null removal.
\Cref{fig:model_scatter} displays the joint distributions of AS and IS across five models. DinoViT attains the best IS/AS trade-off, consistent with its foundation-scale pretraining on a large, diverse corpus beyond ImageNet prior to fine-tuning. This trade-off is evident both in the IS/AS ratio bar plot (panel (b)) and in the orientation of the confidence ellipses in panel~(a). By contrast, ResNext101 shows high AS with substantial variance, which we interpret as class-dependent semantic leakage into its null space. Repeating the comparison with EVA02 \citep{fang2024eva} as the target multimodal space preserves the same model ordering in the ratio analysis (see supplementary material). To further validate the translator, we train classifier heads on principal features before and after translation to CLIP space, obtaining a high Pearson correlation of 0.972 across models (see supplementary material). We also include an extended 12-model sweep as additional coverage across a broader architectural variety.

\subsection{Class analysis}

\begin{figure}[thb]
  \centering
  \begin{subfigure}{0.48\textwidth}
  \includegraphics[width=\textwidth]{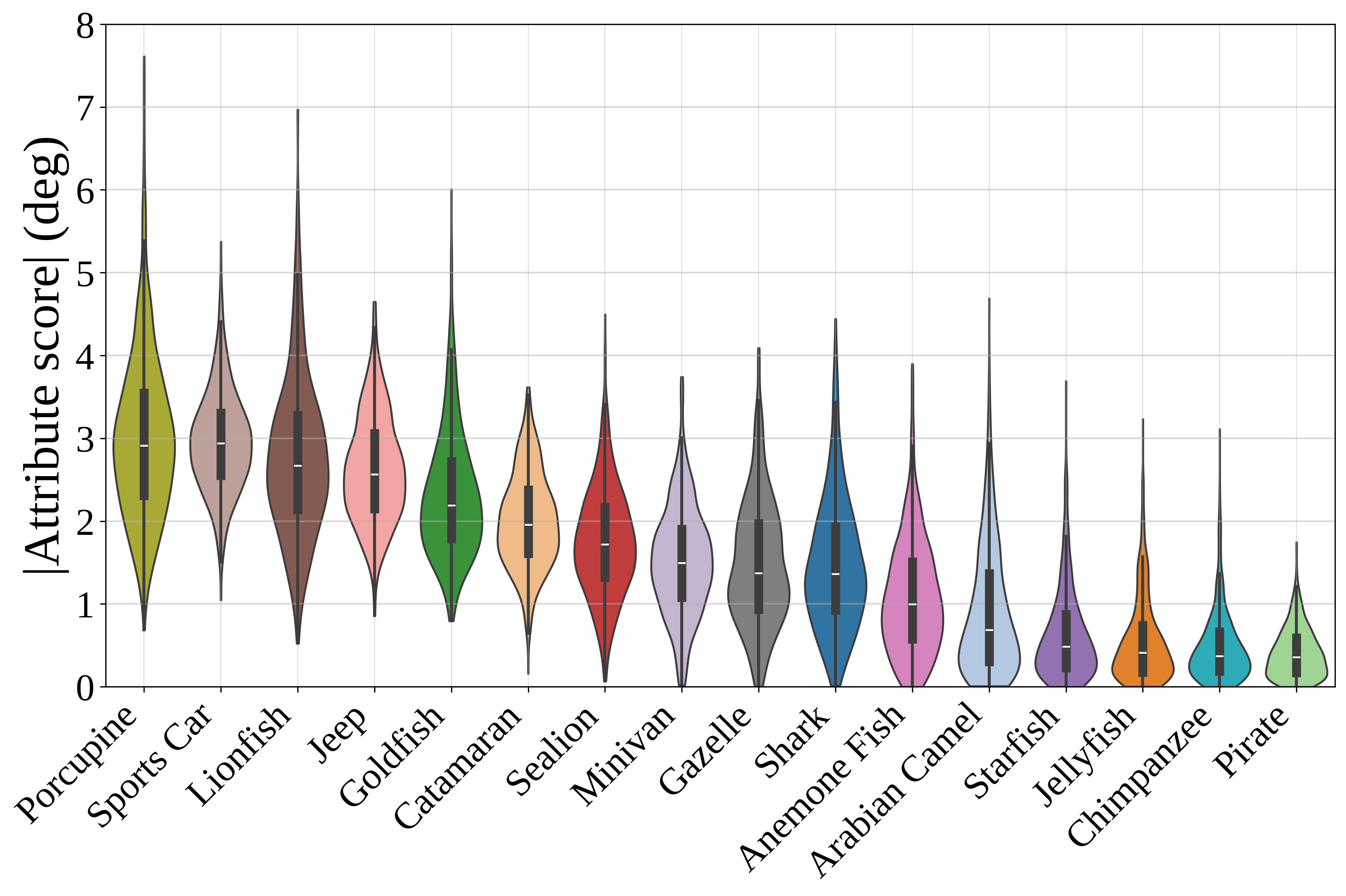}
  \caption{ResNet50}
  \end{subfigure}
  \begin{subfigure}{0.48\textwidth}
  \includegraphics[width=\textwidth]{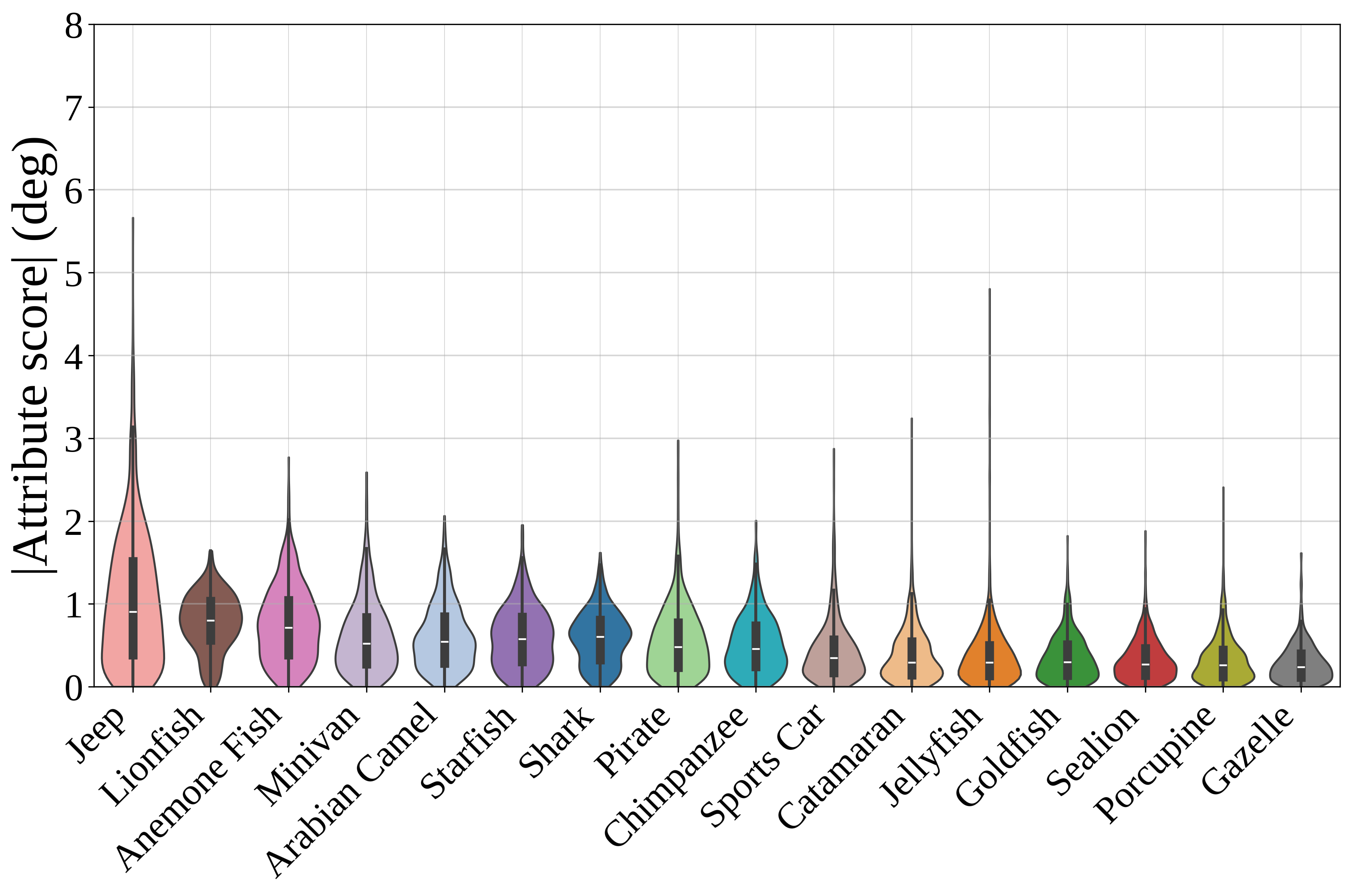}
  \caption{DinoViT}
    \end{subfigure}
    \caption{\textbf{Class Comparison.} DinoViT consistently preserves low semantic leakage across classes, whereas ResNet50 exhibits a pronounced imbalance, with certain classes,
    such as Porcupine and Sports-Car, leaking substantially more semantic information into the null space.}

    \label{fig:violin_between_models}
\end{figure}

We present per class statistics of AS for two of our models, ResNet50 and DinoViT, and report them class by class; see \Cref{fig:violin_between_models}. For each class, AS is measured after null removal. A complete analysis of the other models can be found in the supplementary materials.
DinoViT exhibits stable behavior with very small AS magnitudes (typically $|\mathrm{AS}|<1$), consistent with minimal class-dependent leakage into the null space. By contrast, ResNet50 shows larger and more variable AS across classes. This contrast suggests that DinoViT tends to retain class-relevant semantics within its invariant subspace, whereas ResNet50 appears to possibly rely also on spurious cues, leaving some class-relevant information in the null space. Finally, we observe no significant correlation between the per-class AS rank orderings of the two models, indicating that the effect is model-dependent rather than driven by dataset class structure.

\begin{figure}[t]
  \centering
  \begin{subfigure}{0.48\textwidth}
  \includegraphics[width=\textwidth]{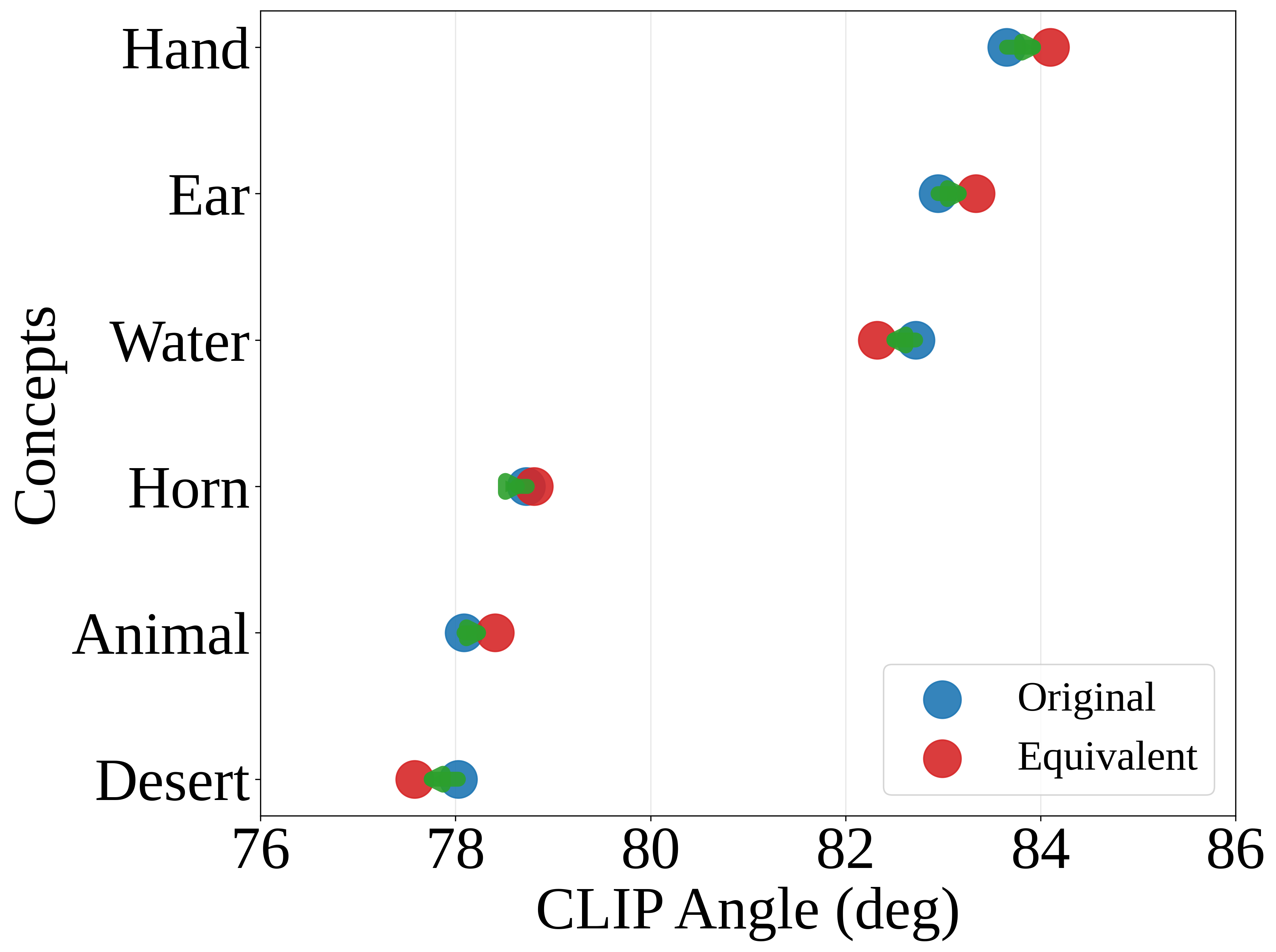}
  \caption{`Arabian Camel` class}
    \end{subfigure}
  \begin{subfigure}{0.48\textwidth}
  \includegraphics[width=\textwidth]{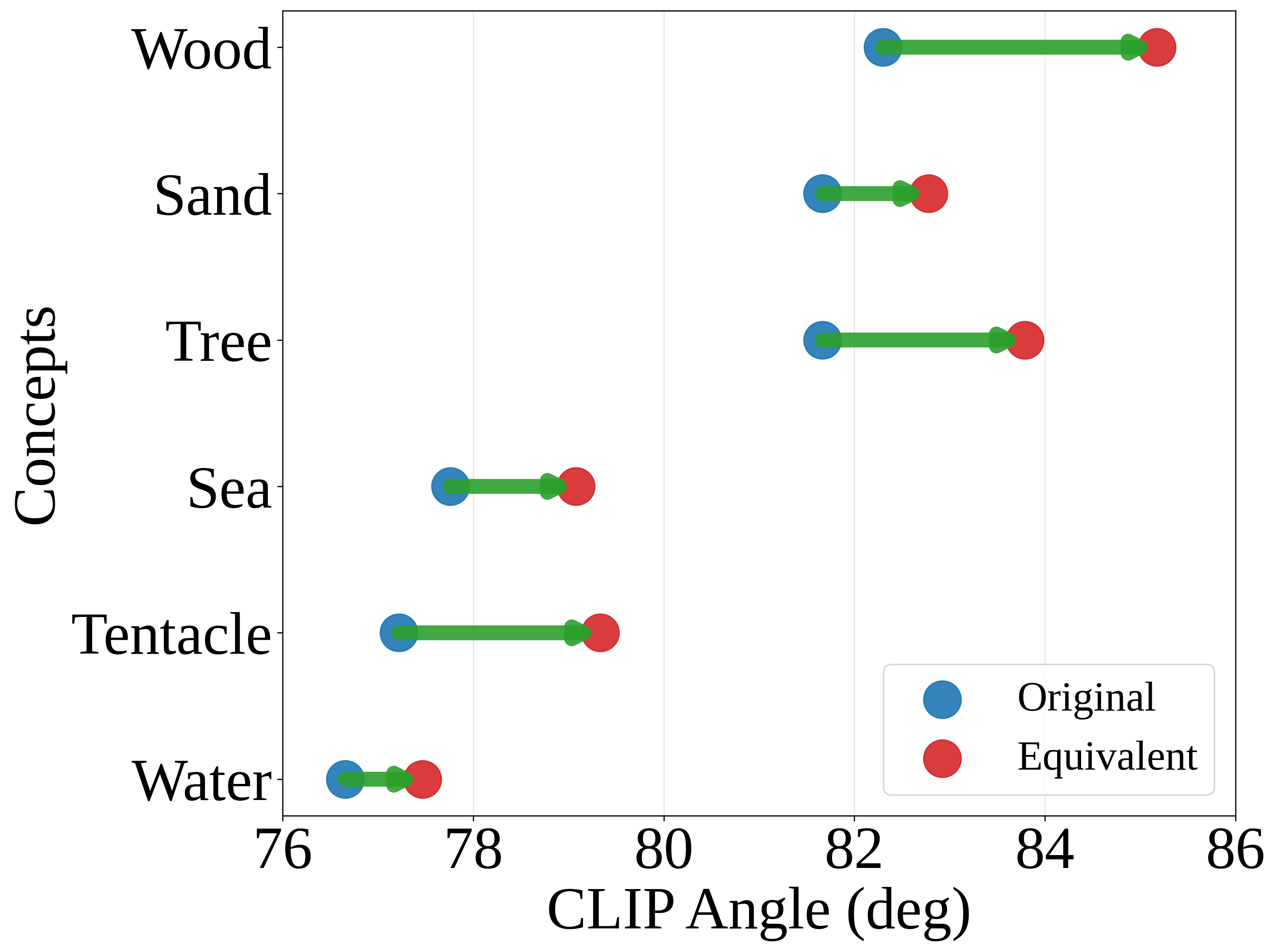}
  \caption{`Jellyfish` class}
   \end{subfigure}
    \caption{\textbf{Open-vocabulary concept analysis.} For DinoViT, we sample $\,\sim\!1300$ images per class and compute the CLIP angle (degrees; lower is more similar) to a set of concepts for (a) ``Arabian Camel'' class and (b) ``Jellyfish'' class. Blue dots denote original features; red dots denote null-removed (equivalent) features. Green arrows connect each pair and represent the Attribute Score after null removal. Longer arrows indicate larger $|\mathrm{AS}|$ (greater class-dependent semantic leakage); shorter arrows indicate minimal leakage.}
    \label{fig:open_voc}
\end{figure}

In \cref{fig:open_voc}, We extend the class analysis to an open vocabulary of concepts. Focusing on DinoViT, we examine two classes, ``Arabian Camel'' and ``Jellyfish''. We measure two quantities: 1) The \emph{angle} between the translated feature and the CLIP concept embedding; 2) the Attribute Score (AS), quantifies how much content related to a concept resides in the null space;
A small AS for loosely related concept can indicate a spurious correlation. Both classes are analyzed through a set contains of 30 concepts, the extreme weakest and strongest are presented. ``Arabian Camel'' features exhibit little to no AS (short green lines), while Desert attains the smallest CLIP angle among the tested concepts. By contrast, ``Jellyfish'' features have substantially larger AS, indicating that concepts are tightly coupled to invariances related to this class in the classifier head. The results on the full set of open-vocabulary concepts and intuition for the scale of AS values is provided in the supplementary materials.


\begin{figure*}[htb]
  \centering
  \includegraphics[width=0.98\textwidth]{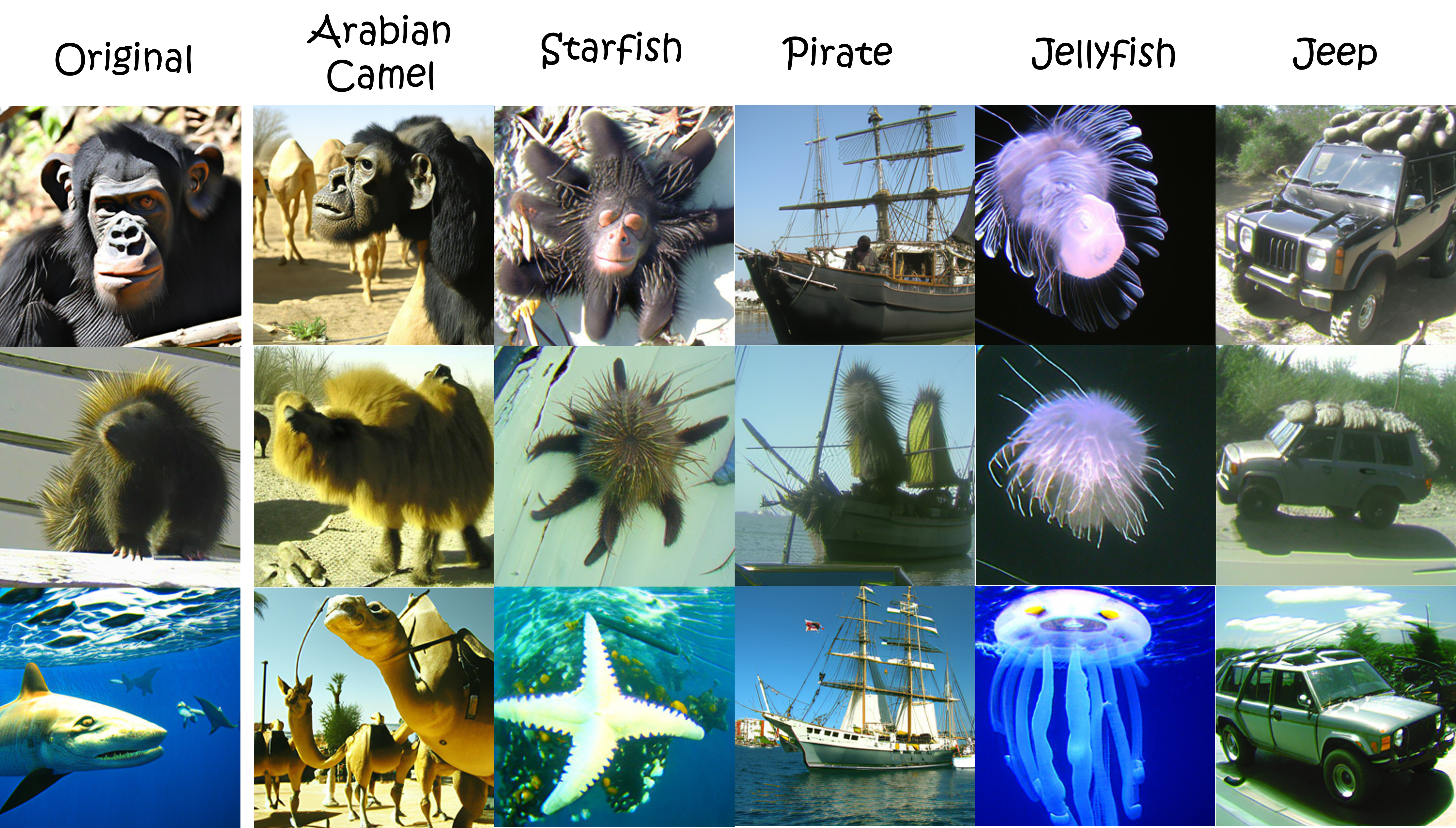}
  \caption{\textbf{Null-space semantic steering (ResNet50).} From each original image (left), we add a small perturbation aligned with the indicated prompt (column headers) but constrained to the classifier head’s null space (projected-gradient direction). Although only the invariant component is modified, the feature’s semantics shift toward the target concepts, illustrating how null-space directions can alter meaning without changing the discriminative subspace.}

  \label{fig:perturb}
\end{figure*}

\subsection{Gradient direction analysis} 

In the previous experiments, we restricted our analysis to equivalent pairs obtained by removing the null component. However, our method supports any null-space direction, including text-conditioned perturbations. In \Cref{fig:perturb}, we illustrate concept-directed perturbations confined to the null space of the ResNet50 classifier head. For each original image (left), we follow the CLIP similarity gradient toward a target prompt, project it onto the null space, and take a step in this direction to obtain an equivalent feature. By construction, the perturbed feature leaves the head logits unchanged. The synthesized renderings, generated with UnCLIP \citep{ramesh2022hierarchical} for visualization, reveal pronounced semantic shifts toward \emph{Arabian Camel}, \emph{Starfish}, \emph{Pirate}, \emph{Jellyfish}, and \emph{Jeep}. This demonstrates the diagnostic value of null-space steering and highlights a security risk: semantics can be manipulated at a single layer while the classifier’s decision remains unaffected.

\Cref{tab:grad_model_comparison} summarizes null-space steps (calibrated to $\mathrm{IS}=40^\circ$) from \emph{Sports Car} toward the prompt ``\texttt{an image of a jellyfish}''. In this setting, DinoViT exhibits low AS, indicating resilience to directed null manipulation. By contrast, EfficientNet and ResNet50 show large AS, suggesting that their null components are easier to steer and that directed invariant perturbations can alter semantics while leaving the logits unchanged.

\begin{table*}[thb]
\centering
\caption{\textbf{Text-gradient null perturbations.} For a fair comparison, each model is perturbed by a fixed null-space step calibrated to $\mathrm{IS}=40^\circ$. We report $|\mathrm{AS}|$ toward the target prompt (mean $\pm$ standard deviation; lower is better). DinoViT attains the lowest value (marked in bold), indicating the greatest resistance to directed null-space manipulation, whereas ResNext101 remains comparatively susceptible.}

\begin{tabular}{l|ccccc}
\toprule
 & ResNet50 & EfficientNet & BiTresnet & DinoViT & ResNext101 \\
 \midrule
 $|\mathrm{AS}|$ towards target & 12.04$\pm$0.25 & 12.38$\pm$0.52 & 9.19$\pm$0.31 & \textbf{5.0$\pm$0.59} & 11.15$\pm$0.53 \\
\bottomrule
\end{tabular}
\label{tab:grad_model_comparison}
\end{table*}

\section{Discussion and Conclusion}
We introduced SING, a novel approach for analyzing invariances in classification networks. Our method systematically generates equivalent images whose logits are, by construction, identical to those of the original image. We demonstrated a wide range of possible analyses: at the model level, SING facilitates fair sensitivity comparisons across architectures; at the class level, it highlights classes that are less robust to semantic shifts; and at the image level, it aids in debugging failure cases.  
SING transforms the null space into measurable and human-readable evidence by constructing equivalent pairs, projecting features into a joint vision-language space, and perturbing only the invariant component. In doing so, it reveals how semantics can drift while logits remain fixed, providing a compact diagnostic that complements accuracy at the levels of models, classes, and individual images.  
Looking ahead, two research directions may help control the null space more directly:  
(i) Directed augmentation during fine-tuning, encouraging small $\mathrm{AS}$ for essential concepts; 
(ii) Linear-algebraic control, using projector regularization, rank adjustment, or constrained updates to move useful semantics from the null space to the principal space while preserving logits.  
SING exposes invariant geometry in a simple, interpretable form, clarifying how semantics can shift while logits remain fixed.

\section*{Acknowledgments}
We would like to acknowledge  
support by the Israel Science Foundation (Grant 1472/23) and by the Ministry of Innovation, Science and Technology (Grant 8801/25).
{
    \small
    \bibliographystyle{CVPR2026/ieeenat_fullname}
    \bibliography{references}
}

\renewcommand{\maketitlesupplementary}{%
  \begin{center}
    \Large\textbf{\thetitle}\\
    \vspace{0.5em}Supplementary Material\\
    \vspace{1.0em}
  \end{center}
}
\onecolumn
\setcounter{page}{1}
\setcounter{section}{0}
\maketitlesupplementary
\addtocontents{toc}{\protect\setcounter{tocdepth}{3}}
{\hypersetup{linkcolor=black}
\tableofcontents
}
\clearpage

\section{Setup and reproducibility}
\subsection{Translator training}
\label{app:Translator_training}

As described in the paper, each translator trained on a specific classifier and its task is to map features from the penultimate layer $f \in \mathbb{R}^{d}$ to a CLIP image feature $e \in \mathbb{R}^{d_e}$. Nonlinear translators were trained directly in PyTorch \citep{paszke2019pytorch}, while linear translators were fitted by ridge regression using scikit-learn \citep{scikit-learn} and then ported to PyTorch for unified inference. The hyperparameters were chosen using sweeps logged in Weights \& Biases \citep{wandb}.
We compared three training objectives: \newline
\begin{enumerate}

    \item Mean squared error (MSE) loss:
        \begin{equation}
        \label{eq:translator_mse}
        \mathcal{L}_{\mathrm{MSE}}(f,e)
        = \left\lVert T_{\theta}(f) - e \right\rVert_2^2.
        \end{equation}
    \item Cosine similarity loss:
        \begin{equation}
        \mathcal{L}_{\mathrm{cos}}(f,e)
        = 1 -
        \frac{T_{\theta}(f) \cdot e}
             {\left\lVert T_{\theta}(f) \right\rVert_2
              \left\lVert e \right\rVert_2}.
        \label{eq:translator_cos}
        \end{equation}
    \item MSE + Cosine loss
    \newline
\end{enumerate}

\begin{figure}[h]
  \centering
  \begin{subfigure}{0.32\textwidth}
    \includegraphics[width=\textwidth]{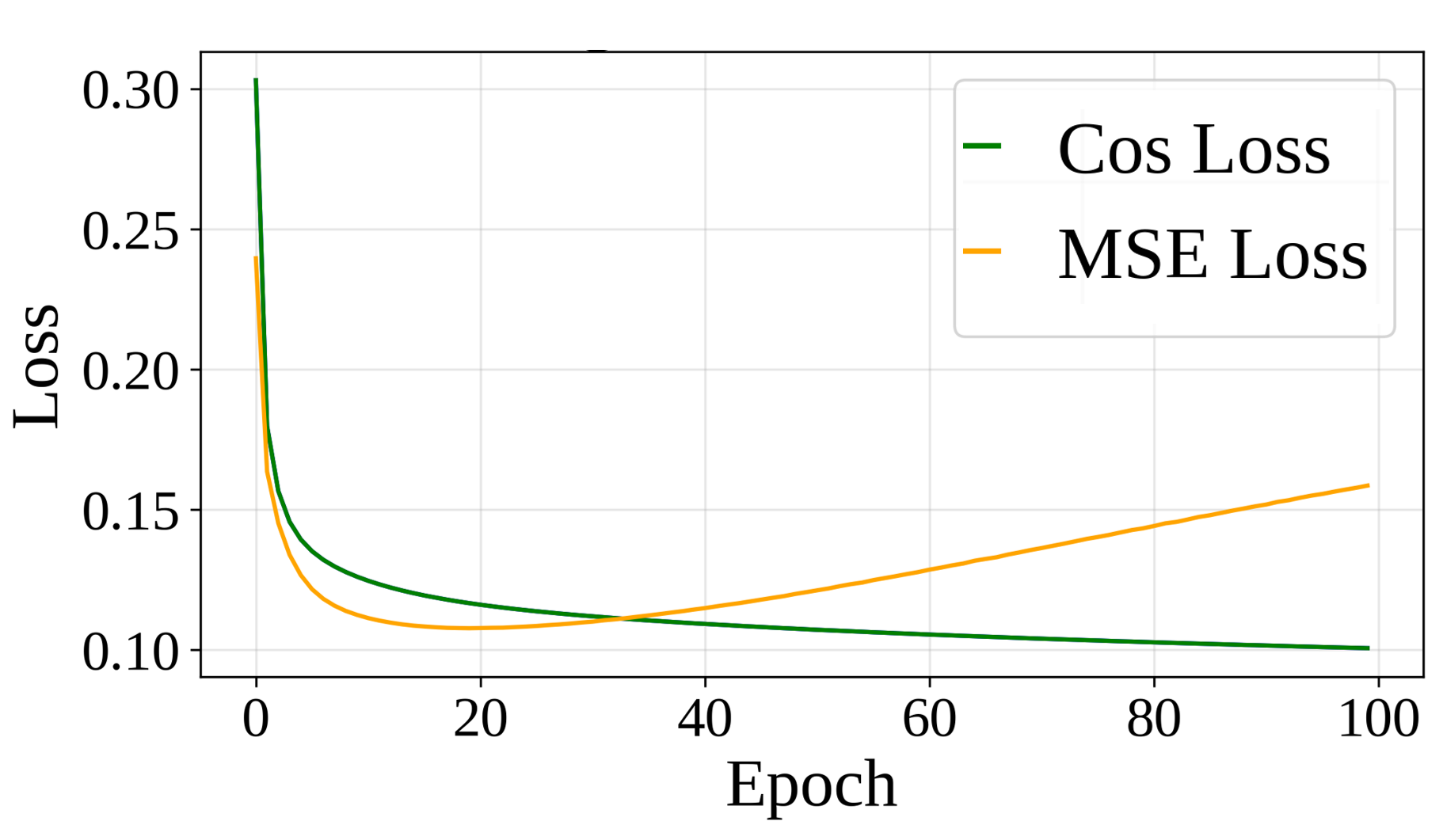}
    \caption{Cosine-only loss $\mathcal{L}_{\mathrm{cos}}$.}
  \end{subfigure}
  \begin{subfigure}{0.32\textwidth}
    \includegraphics[width=\textwidth]{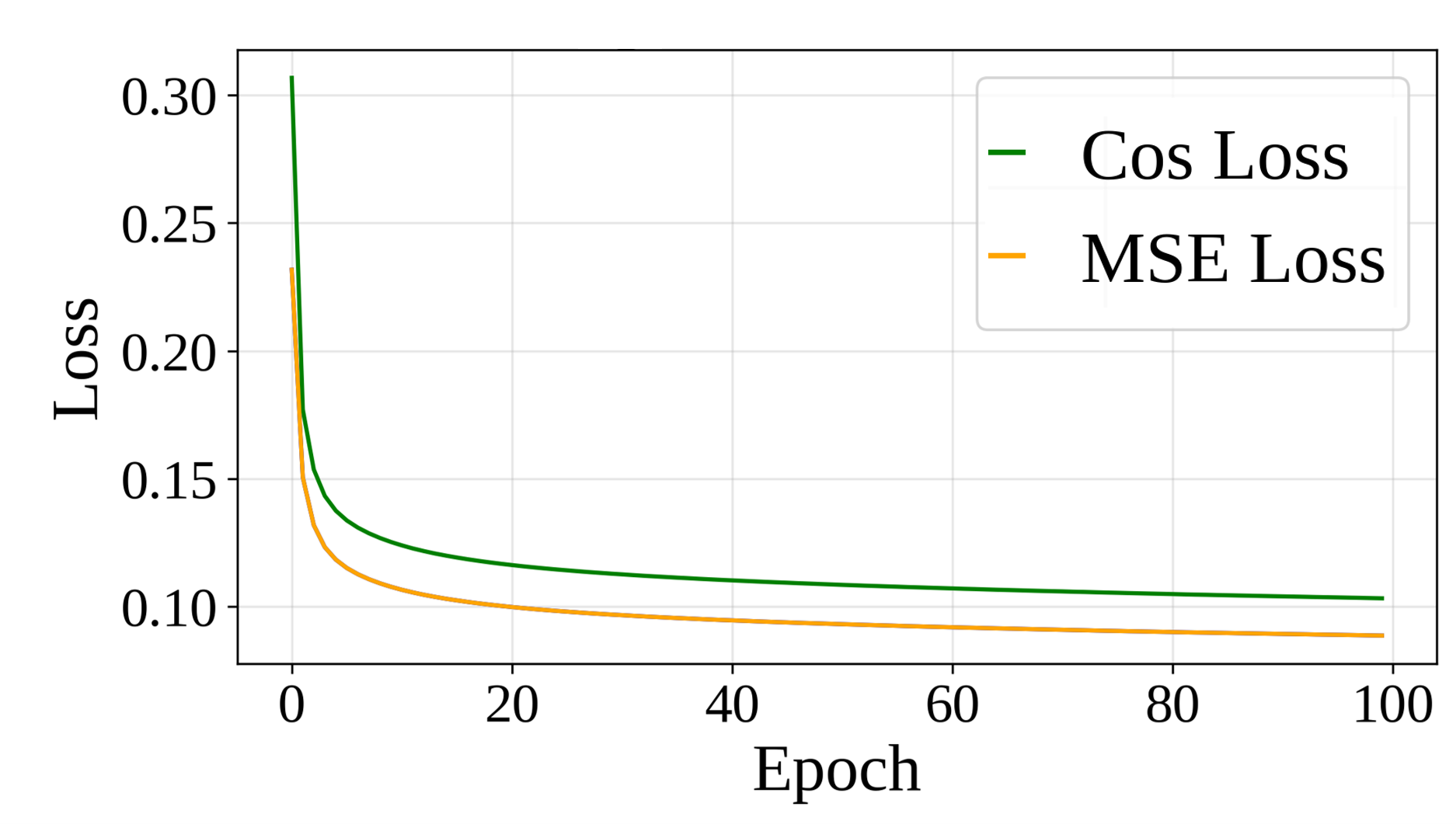}
    \caption{MSE-only loss $\mathcal{L}_{\mathrm{MSE}}$.}
  \end{subfigure}
  \begin{subfigure}{0.32\textwidth}
    \includegraphics[width=\textwidth]{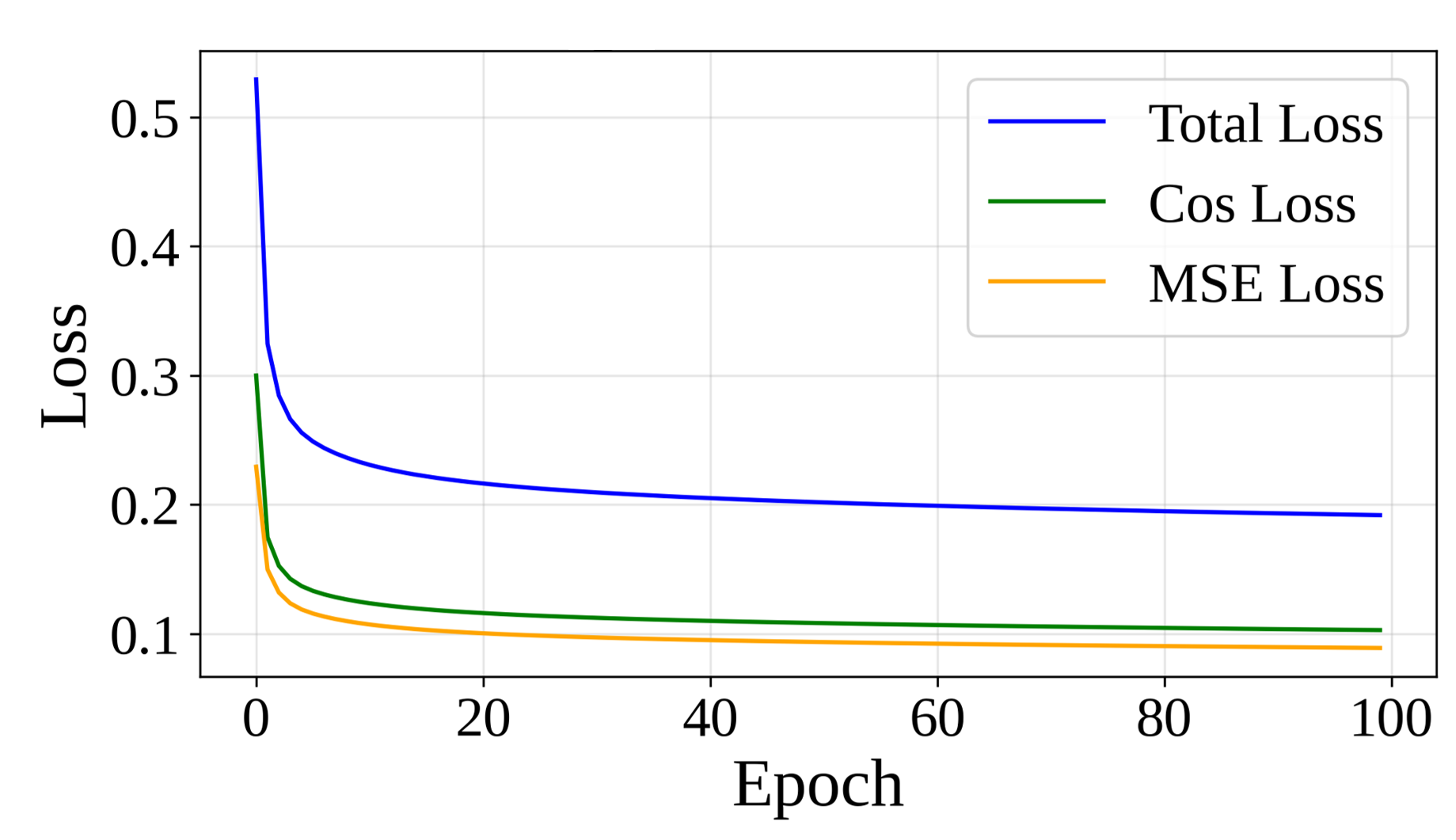}
    \caption{Joint loss $\mathcal{L}_{\mathrm{joint}}$.}
  \end{subfigure}
  \caption{Training losses for the different translator objectives.
  Minimizing the MSE loss also improves cosine similarity, whereas
  cosine-only training leaves the MSE substantially higher.}
  \label{fig:loss_plots}
\end{figure}

For all three cases, we applied $L_2$ regularization.
In practice, minimizing $\mathcal{L}_{\mathrm{MSE}}$ alone proved sufficient to achieve high cosine similarity, whereas optimizing $\mathcal{L}_{\mathrm{cos}}$ alone does not reliably reduce MSE, suggesting an asymmetric relationship between the two objectives. This trend is illustrated in \Cref{fig:loss_plots}.

Our baseline translator is a linear map chosen for stability.
To compare linear and non-linear translators, we evaluate three additional.
As for Nonlinear architectures, we tried the following combinations:\newline
\begin{enumerate}
    \item A 3-layer MLP with blocks of the form
        LayerNorm–GELU–Dropout-FC
        \citep{ba2016layernorm,hendrycks2016gelu,srivastava2014dropout}
    \item A 4-layer MLP with the same block.
    \item A residual MLP with one residual blocks and one projection    layer.
    \newline
\end{enumerate}
All nonlinear translators were optimized with AdamW \citep{loshchilov2019decoupled}.
with learning rate $1 \times 10^{-4}$ and weight decay $\lambda = 0.1$.

\begin{figure}[h]
  \centering
  \includegraphics[width=0.48\textwidth]{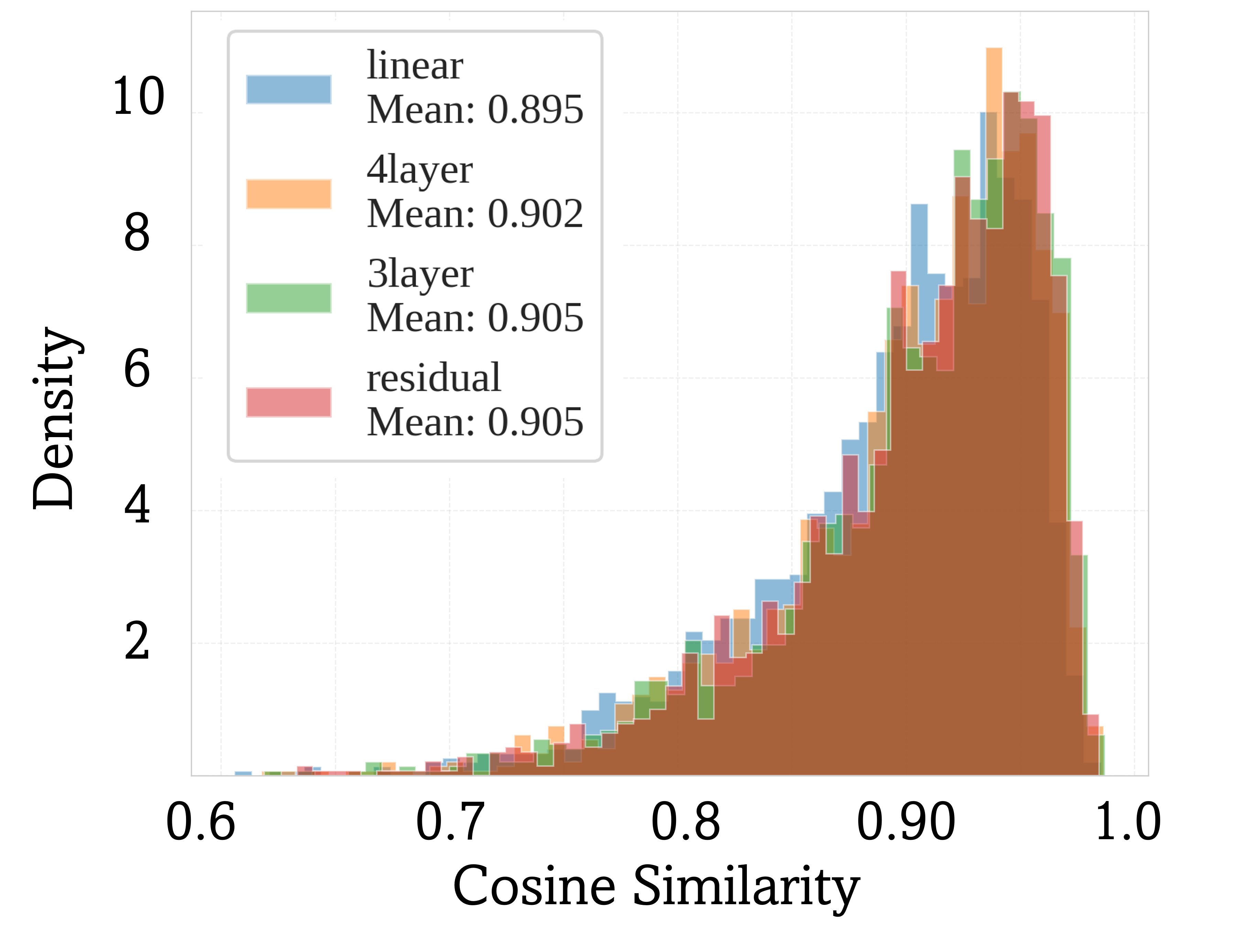}
  \caption{Cosine similarity distribution of different architectures between the translated and the original CLIP features, over ~2k ImageNet features from 16 classes. All the histograms are leaned towards high correlation}
  \label{fig:cosine_dist}
\end{figure}

We report validation results over a 2,000-image subset in \Cref{tab:architecture_metrics} and \Cref{fig:cosine_dist} showing no significant advantage of any non-linear variant over the linear translator.

\begin{table}[b]
    \centering
    \caption{Validation results for different translator architectures on a
    2\,000-image validation subset from 16 classes.
    None of the non-linear architectures shows a significant advantage over
    the linear translator.}
    \begin{tabular}{lcc}
        \toprule
        Architecture & Mean cosine similarity & Validation MSE \\
        \midrule
        3-layer MLP  & 0.9049 & 0.082246 \\
        Residual MLP & 0.9045 & 0.082366 \\
        4-layer MLP  & 0.9023 & 0.084346 \\
        Linear       & 0.8946 & 0.091355 \\
        \bottomrule
    \end{tabular}
    \label{tab:architecture_metrics}
\end{table}

\section{Null space validation}
\label{app:null_validation}

Let $f$ denote the penultimate classifier feature and
$\ell(f) \in \mathbb{R}^{C}$ the corresponding vector of logits for $C$
classes.
We define the logit change induced by a perturbation $\delta$ as
\begin{equation}
\Delta_{\ell}(f,\delta)
= \left\lVert \ell(f+\delta) - \ell(f) \right\rVert_2 .
\label{eq:logit_change_metric}
\end{equation}

\begin{figure}[thb]
  \centering
  \begin{subfigure}{0.48\textwidth}
    \includegraphics[width=\textwidth]{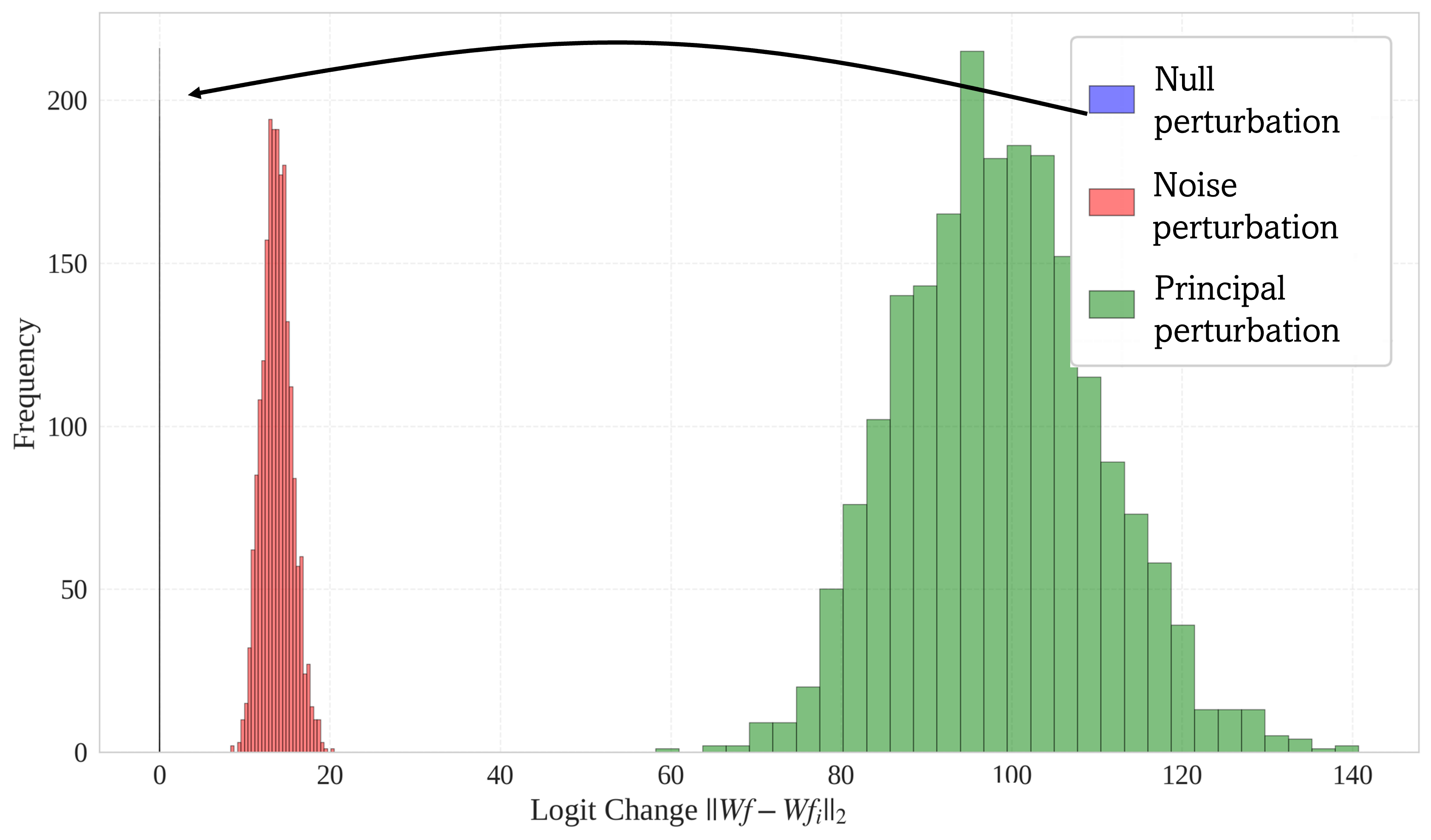}
    \caption{Logit change under null vs.\ random perturbations.}
  \end{subfigure}
  \begin{subfigure}{0.48\textwidth}
    \includegraphics[width=\textwidth]{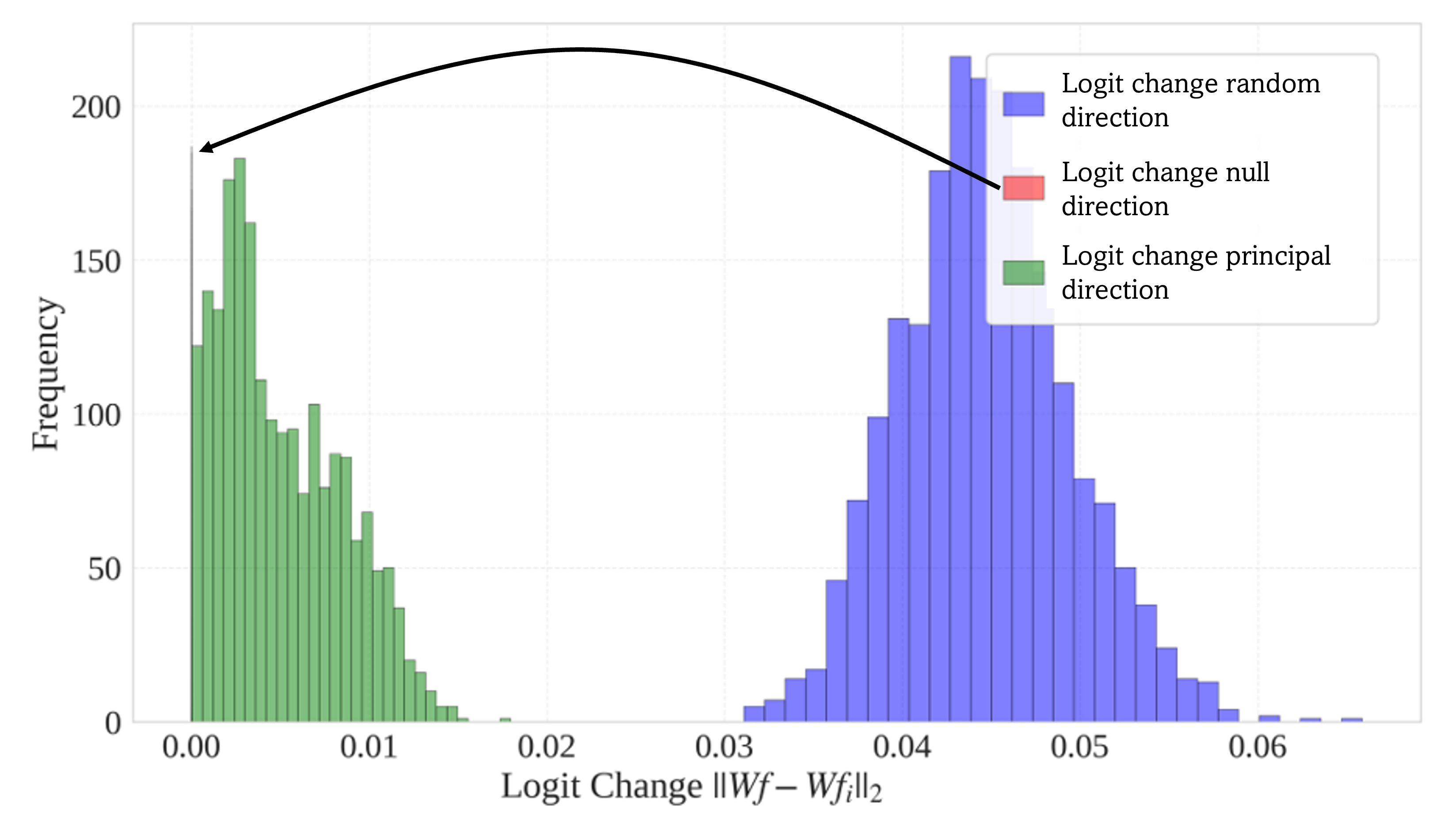}
    \caption{Logit change under principal vs.\ random perturbations.}
  \end{subfigure}
  \caption{Distribution of logit changes
  $\Delta_{\ell}(f,\delta)$ for null-space, random, and principal
  perturbations.
  Null-space perturbations leave logits almost unchanged, whereas principal
  perturbations induce large logit shifts.}
  \label{fig:logit_change}
\end{figure}

We compare three types of perturbations with matched $\ell_2$-norm:
(i) a null perturbation $\delta_{\mathrm{null}}$ in the approximate null space
of the classifier head, satisfying
\begin{equation}
W \, \delta_{\mathrm{null}} \approx 0 ,
\label{eq:null_def}
\end{equation}
where $W$ are the head weights;
(ii) a random perturbation $\delta_{\mathrm{rand}}$ sampled from an
isotropic Gaussian and rescaled to the same norm; and
(iii) a principal perturbation $\delta_{\mathrm{principal}}$ chosen along a
direction that strongly affects the logits (e.g.\ a leading sensitive
direction for the predicted class) rescaled as well to the null perturbation magnitude.
For each type we compute the logit change in L2-norm over a validation set and
summarize the distribution in \Cref{fig:logit_change}.

\begin{figure*}[b]
  \centering
  \includegraphics[width=0.78\textwidth]{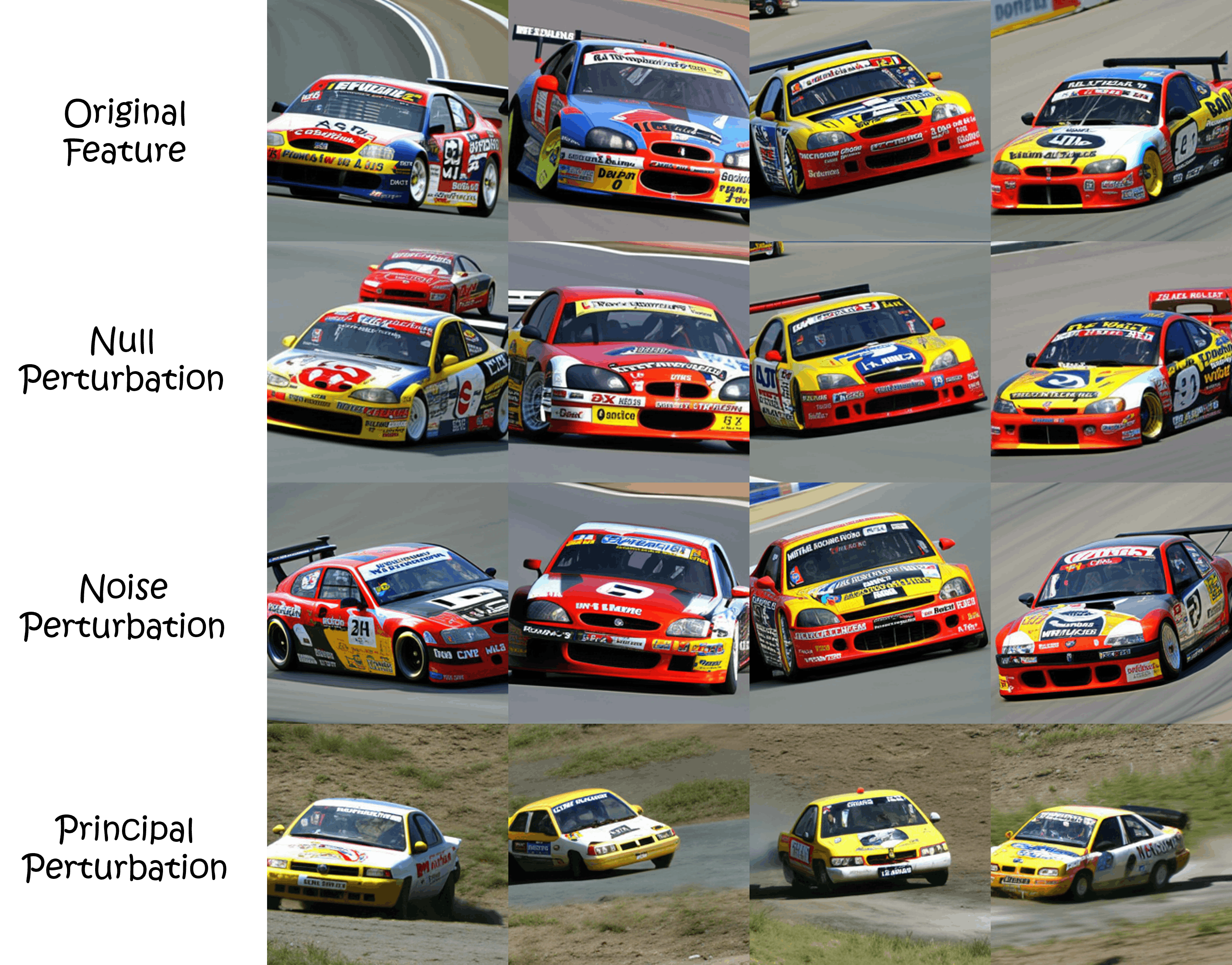}
  \caption{UnCLIP generations of a single feature under three perturbation
  types (null, random, principal) across four random seeds.
  Null-space perturbations preserve the global class semantics, while random
  and principal perturbations produce more noticeable semantic changes.}
  \label{fig:pert_example}
\end{figure*}

As expected, null-space perturbations produce negligible logit changes,
while random and principal perturbations have a noticeable shifts.
In \Cref{fig:pert_example} we illustrate the corresponding UnCLIP
generations for a single feature under these three perturbations and
multiple seeds.

\section{Image-level and visualization details}
\subsection{Angle visual interpretation}
\label{app:sem_interp}
\begin{figure}
\centering
\begin{minipage}{0.4\linewidth}
  \centering
  \includegraphics[width=\linewidth]{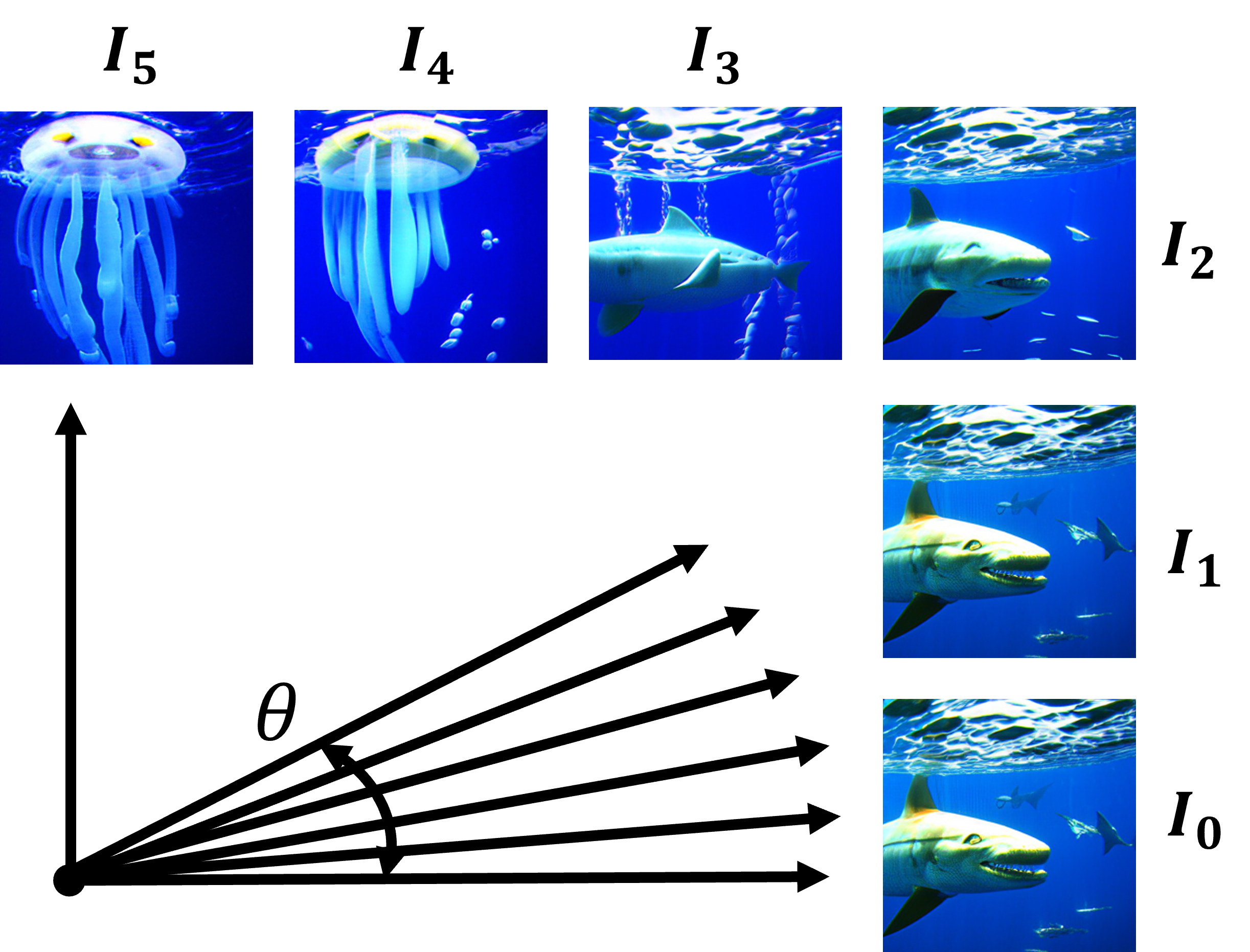}
\end{minipage}\hfill
\begin{minipage}{0.4\linewidth}
  \centering
  \begin{tabular}{lrr}
    \toprule
    Image & AS ($^\circ$) & IS ($^\circ$) \\
    \midrule
    $I_0$ & 0.00 & 0.0 \\
    $I_1$ & 1.58 & 4.0 \\
    $I_2$ & 3.80 & 10.8 \\
    $I_3$ & 4.70 & 23.0 \\
    $I_4$ & 9.48 & 29.0 \\
    $I_5$ & 11.29 & 36.2 \\
    \bottomrule
  \end{tabular}
\end{minipage}

\caption{Example images with different attribute scores (AS) and image
scores (IS), illustrating the relationship between angular distance and
perceived semantic change.
Small angles correspond to nearly identical images, while larger angles
reflect more significant semantic changes.}
\label{tab:angles}
\end{figure}

For the readers convenience, we provide a visual interpretation of the angles we measured along the paper. For two non-zero vectors $u$ and $v$ we define the angle in degrees
\begin{equation}
\theta(u,v) =
\arccos\left(
\frac{u \cdot v}{\lVert u \rVert_2 \, \lVert v \rVert_2}
\right) \cdot \frac{180}{\pi}.
\label{eq:angle_def}
\end{equation}
The attribute score (AS) and image score (IS) used in the main paper are instances of $\theta(\cdot,\cdot)$ applied in CLIP image-embedding space. \Cref{tab:angles} provides a concrete mapping between AS/IS values and visual changes for a single example.
Angles below roughly $3^\circ$ in AS and $10^\circ$ in IS correspond to
barely perceptible changes, while larger angles produce clear semantic
differences such as pose or shape variations.

\subsection{Visualization with UnCLIP}
\label{app:unclip}

UnCLIP is a two-stage image generator: a \emph{prior} maps text to a CLIP
image embedding, and a diffusion-based \emph{decoder} with super-resolution
modules synthesizes the corresponding image \citep{ramesh2022hierarchical}.
CLIP encoders normalize image and text embeddings to unit length and compare
them using cosine similarity, so semantic information is primarily encoded
in the angular component on the unit hypersphere \citep{radford2021clip}.

We use trained translators $T_{\Theta}$ to map classifier features $f$ and
their perturbed variants $\tilde{f}$ into the CLIP image-embedding space.
Given a feature and its equivalent feature set translated to CLIP,
$T_\Theta(f)$ and $T_\Theta(\tilde{f})$, we rescale the translated
equivalent feature to match the norm of the original:
\begin{equation}
\hat{T}_\Theta(\tilde{f}) =
T_\Theta(\tilde{f}) \,
\frac{\left\lVert T_\Theta(f) \right\rVert_2}
     {\left\lVert T_\Theta(\tilde{f}) \right\rVert_2}.
\label{eq:unclip_rescale}
\end{equation}
This preserves the angular relationships while restoring the radial
component, preventing distortions in the visualizations due to radial drift.

To ensure that observed visual differences are solely attributable to
changes in the classifier feature $f$, we remove the stochasticity in the
diffusion sampling process.
We fix the random seed, draw a single Gaussian noise tensor with
\texttt{randn\_tensor}, scale it by the scheduler's
\texttt{init\_noise\_sigma}, and reuse this tensor for all images in the
batch and for both the decoder and super-resolution stages.
For a fixed CLIP image embedding, this procedure yields deterministic
outputs.

\begin{figure}
  \centering
  \includegraphics[width=0.68\textwidth]{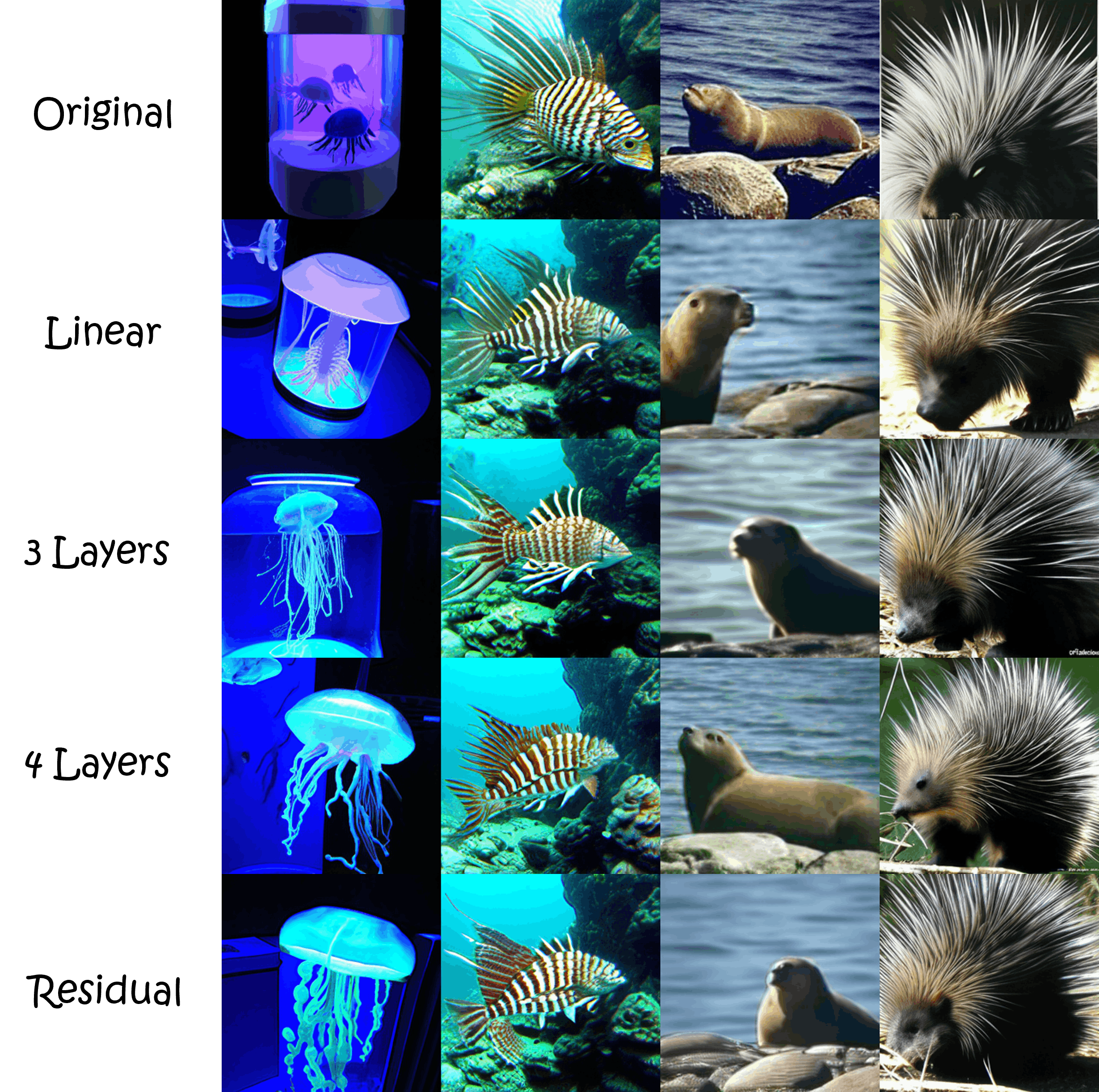}
  \caption{UnCLIP generations from a single classifier feature translated
  by different translator architectures.
  Despite small quantitative differences in cosine similarity, the
  resulting visualizations are qualitatively consistent.}
  \label{fig:gen_example}
\end{figure}

Our implementation uses the \textbf{Karlo-v1.0.alpha} UnCLIP model
\citep{kakaobrain2022karlo-v1-alpha}, which follows the original OpenAI
framework \citep{ramesh2022hierarchical}.
The system includes frozen CLIP text and image encoders, a projection
layer into the decoder space, a \texttt{UNet2DConditionModel} decoder,
two \texttt{UNet2DModel} super-resolution networks, and
\texttt{UnCLIPScheduler} instances for both stages.
A generation example of the same feature translated by different
translators is shown in \Cref{fig:gen_example}.

\section{Model-level result extensions}
\label{app:additional_model_level_validations}
\begin{figure}
  \centering
  \begin{subfigure}{0.32\textwidth}
    \includegraphics[width=\textwidth]{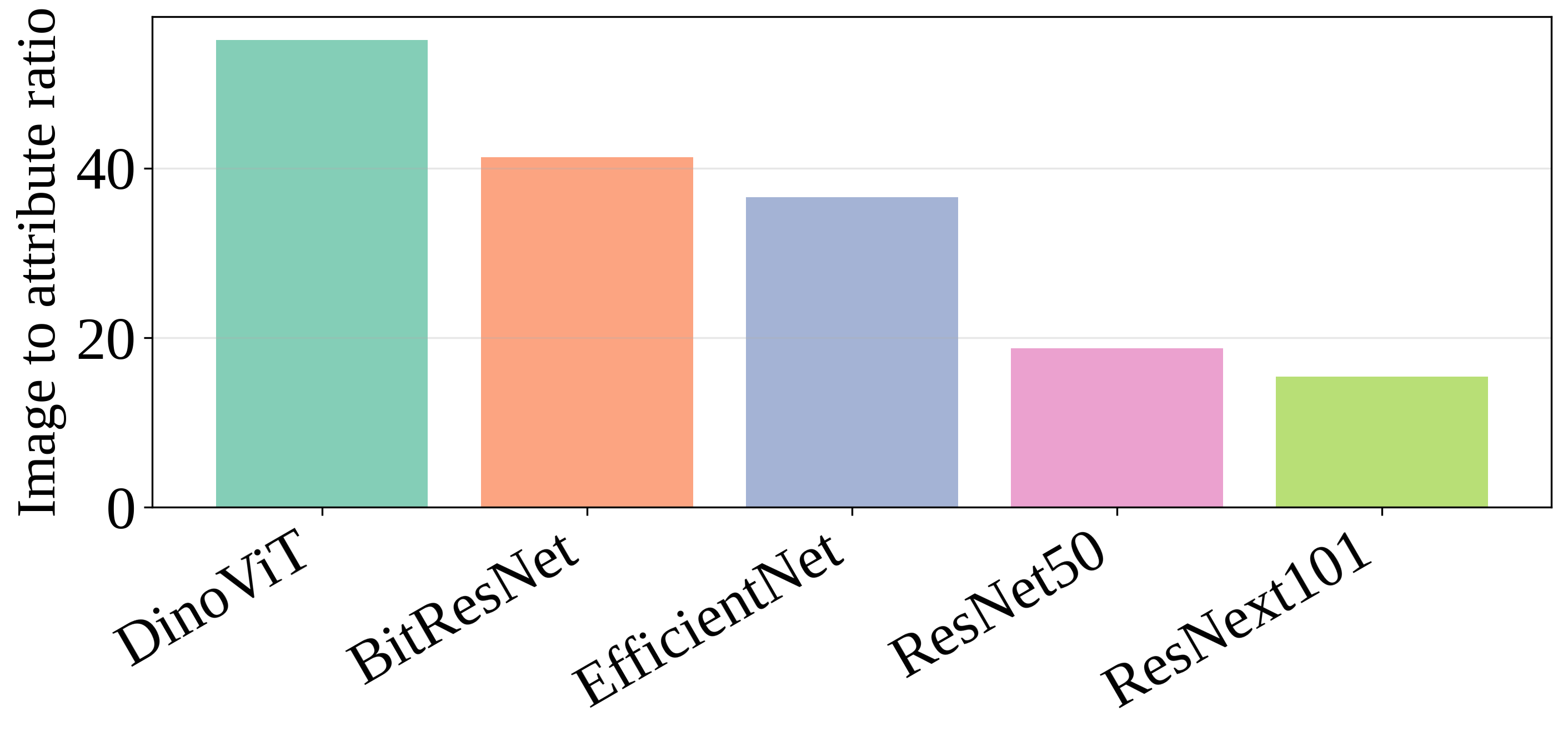}
    \caption{5-model ratio comparison in EVA02 space.}
  \end{subfigure}
  \begin{subfigure}{0.32\textwidth}
    \includegraphics[width=\textwidth]{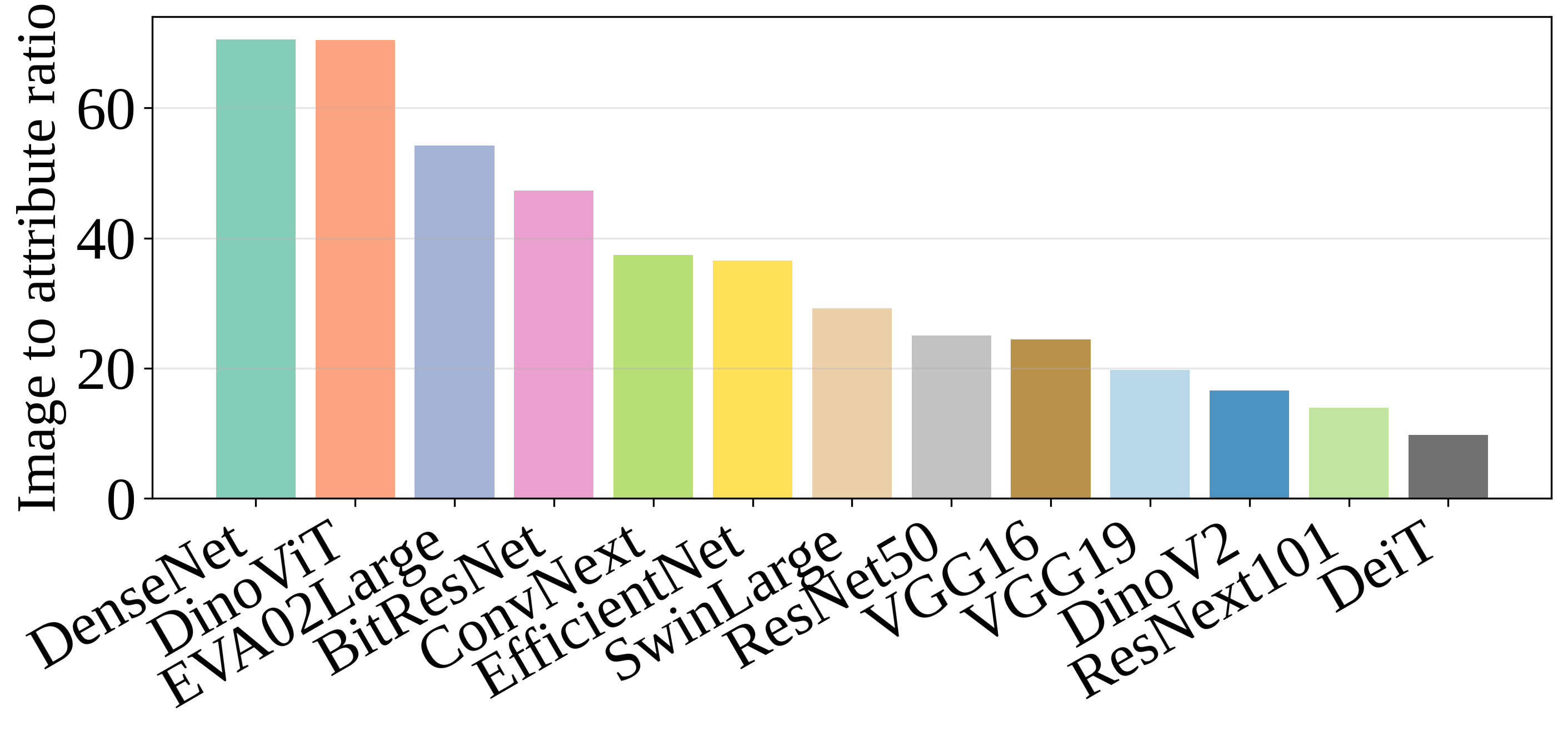}
    \caption{Extended ratio comparison over 12 models.}
  \end{subfigure}
  \begin{subfigure}{0.32\textwidth}
    \includegraphics[width=\textwidth]{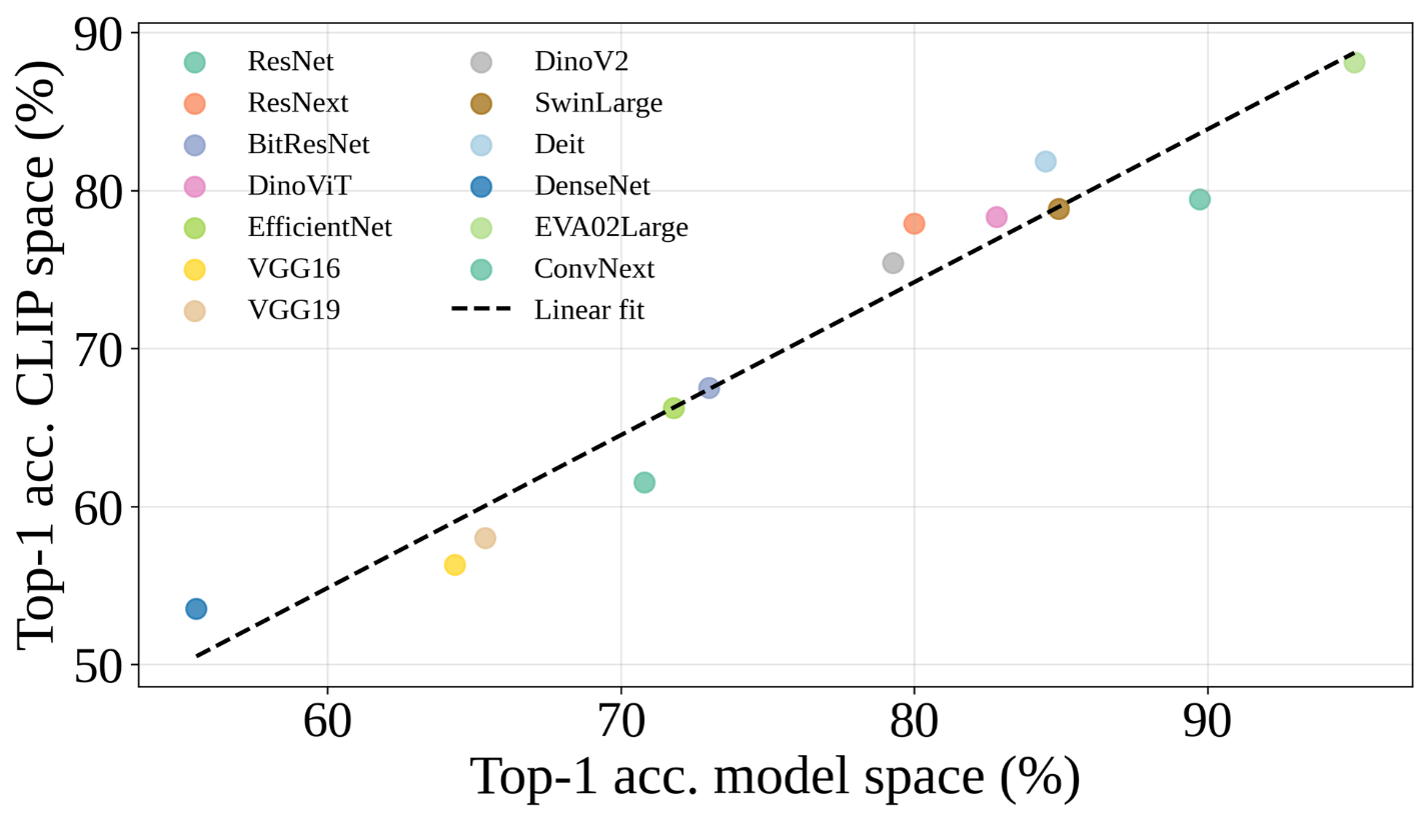}
    \caption{Translator robustness on principal features (Pearson 0.972).}
  \end{subfigure}
  \caption{Additional model-level validations used in the camera-ready update.}
  \label{fig:additional_model_level_validations}
\end{figure}

We include three additional checks requested during rebuttal integration.
First, we repeat the 5-model ratio comparison with EVA02 as the target multimodal space.
Second, we expand the ratio comparison from 5 models to 13 models pretrained on ImageNet \citep{deng2009imagenet} to increase architectural variety. the list of all models can be found in \ref{tab:models}.
Third, we evaluate translator robustness by training classifier heads on 500k principal features before and after translation to CLIP space, and computing the model-wise Pearson correlation of classification accuracy. \ref{fig:additional_model_level_validations}.
The resulting Pearson score is 0.972, indicating strong consistency between the original-principal and translated-principal feature spaces.
\begin{table}[htb]
    \centering
    \caption[Models list]{List of models from CNNs to ViTs that we used as our test subjects.}
    \begin{tabular}{|l|c|}
    \hline
        Model & ImageNet Top-1 Acc (\%) \\ \hline
        VGG-16 \cite{simonyan2014very} & 71.6 \\
        VGG-19 \cite{simonyan2014very} & 72.4 \\
        DenseNet-121 \cite{huang2017densely} & 74.4 \\
        ResNet50 \cite{he2016deep} & 76.1 \\
        DinoViT \cite{caron2021emerging} & 84.0 \\
        EfficientNet-B0 (NS) \cite{xie2020self} & 78.7 \\
        BiT-ResNet (M-R50x1) \cite{kolesnikov2020big} & 80.4 \\
        ResNeXt-101 32x8d (WSL) \cite{mahajan2018exploring} & 82.6 \\
        ConvNeXt-Base \cite{liu2022convnet} & 85.8 \\
        Swin-L \cite{liu2021swin} & 86.3 \\
        DINOv2-L \cite{oquab2023dinov2} & 86.5 \\
        DeiT-3-L/16 \cite{touvron2022deit} & 87.7 \\
        EVA-02-L \cite{fang2023eva} & 89.9 \\ \hline
    \end{tabular}
    \label{tab:models}
\end{table}


\section{Class-level analyses}
\label{app:class_level_analyses}

\begin{figure}[h]
  \centering
  \begin{subfigure}[t]{0.48\textwidth}
    \centering
    \includegraphics[width=\textwidth]{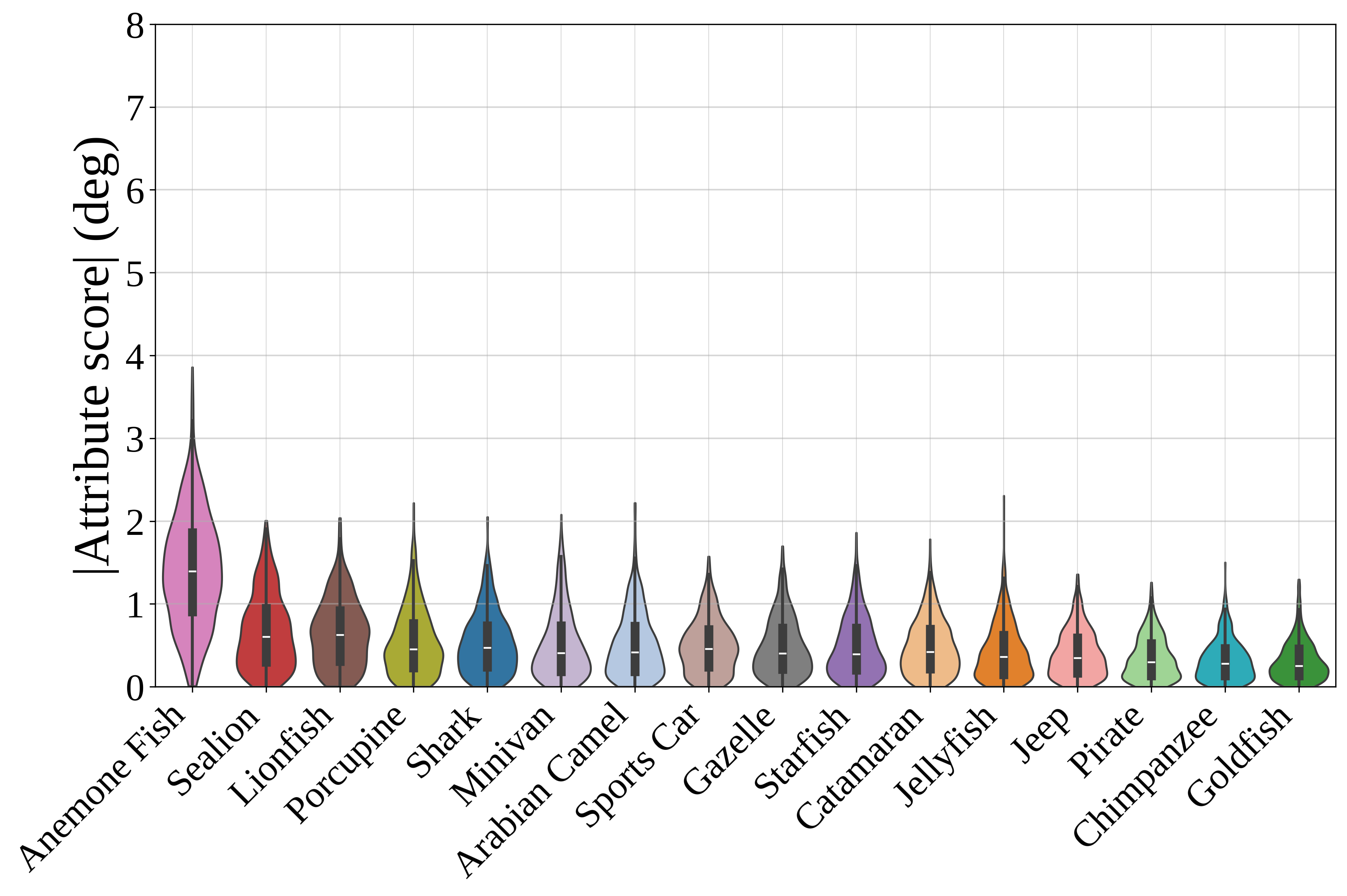}
    \caption{BiT-ResNet \citep{kolesnikov2020big}.}
    \label{fig:bitresnet_violin}
  \end{subfigure}
  \hfill
  \begin{subfigure}[t]{0.48\textwidth}
    \centering
    \includegraphics[width=\textwidth]{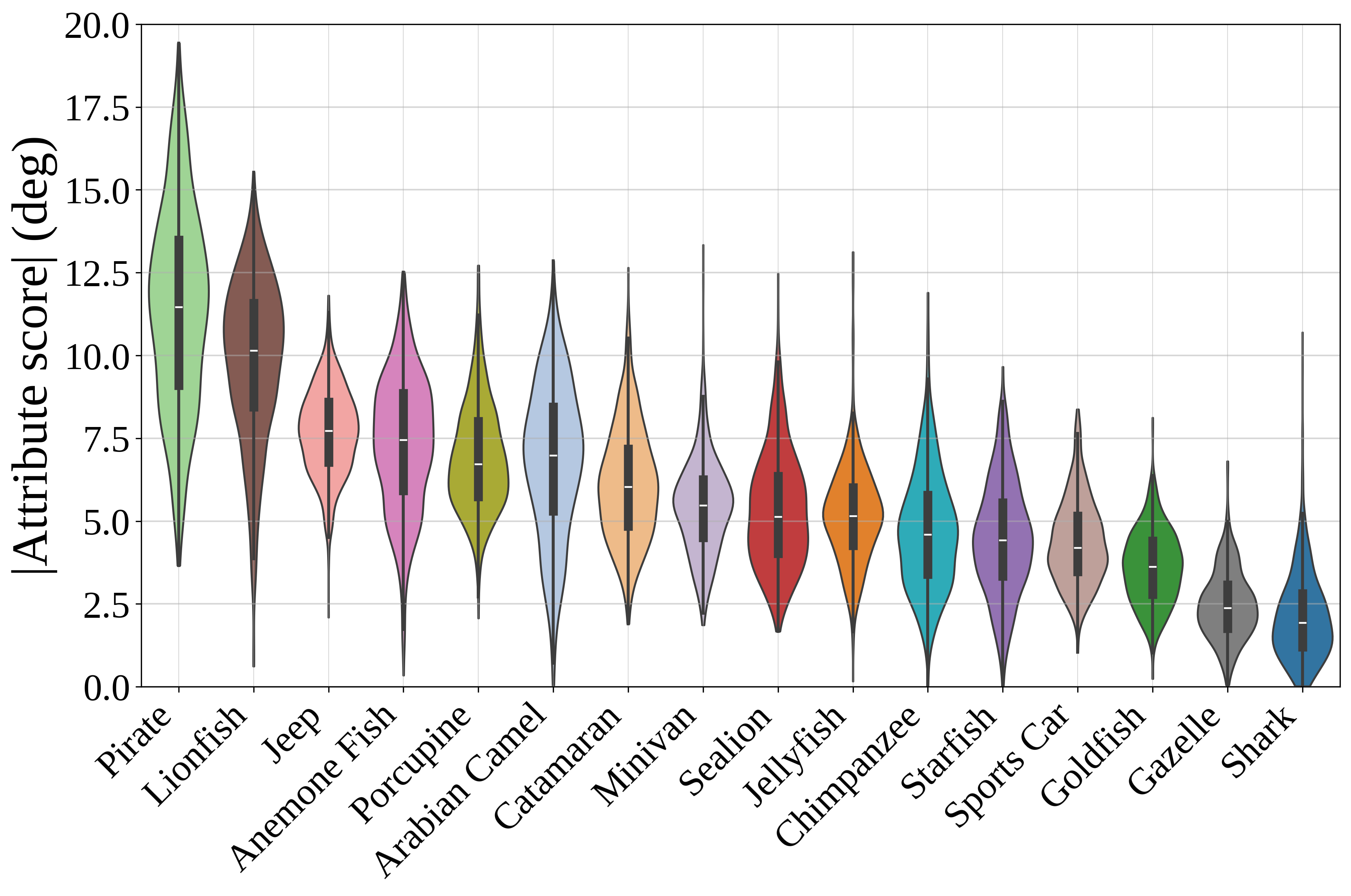}
    \caption{ResNeXt \citep{mahajan2018exploring}.}
    \label{fig:resnext_violin}
  \end{subfigure}

  \vspace{1em}

  \begin{subfigure}[t]{0.48\textwidth}
    \centering
    \includegraphics[width=\textwidth]{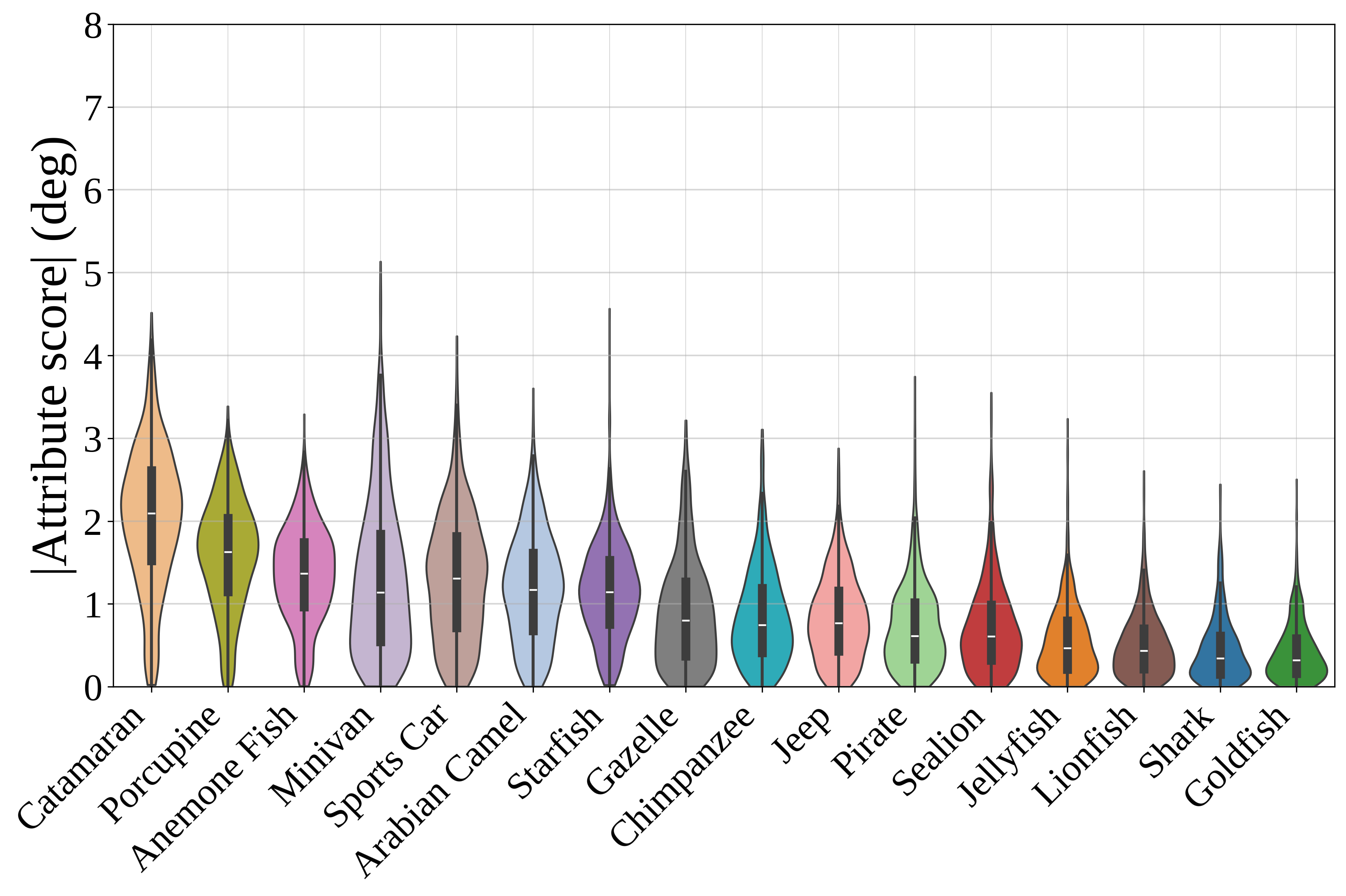}
    \caption{EfficientNet \citep{xie2020self}.}
    \label{fig:efficientnet_violin}
  \end{subfigure}

  \caption{Per-class distribution of null-space semantic angle changes across
  three architectures. Each violin corresponds to a single class; narrow
  distributions around zero indicate classes largely invariant to null-space
  perturbations. The consistent pattern across architectures with different
  inductive biases confirms the generality of our observations.}
  \label{fig:violin_all}
\end{figure}

\begin{figure}[tbh]
  \centering
  \includegraphics[width=0.85\textwidth]{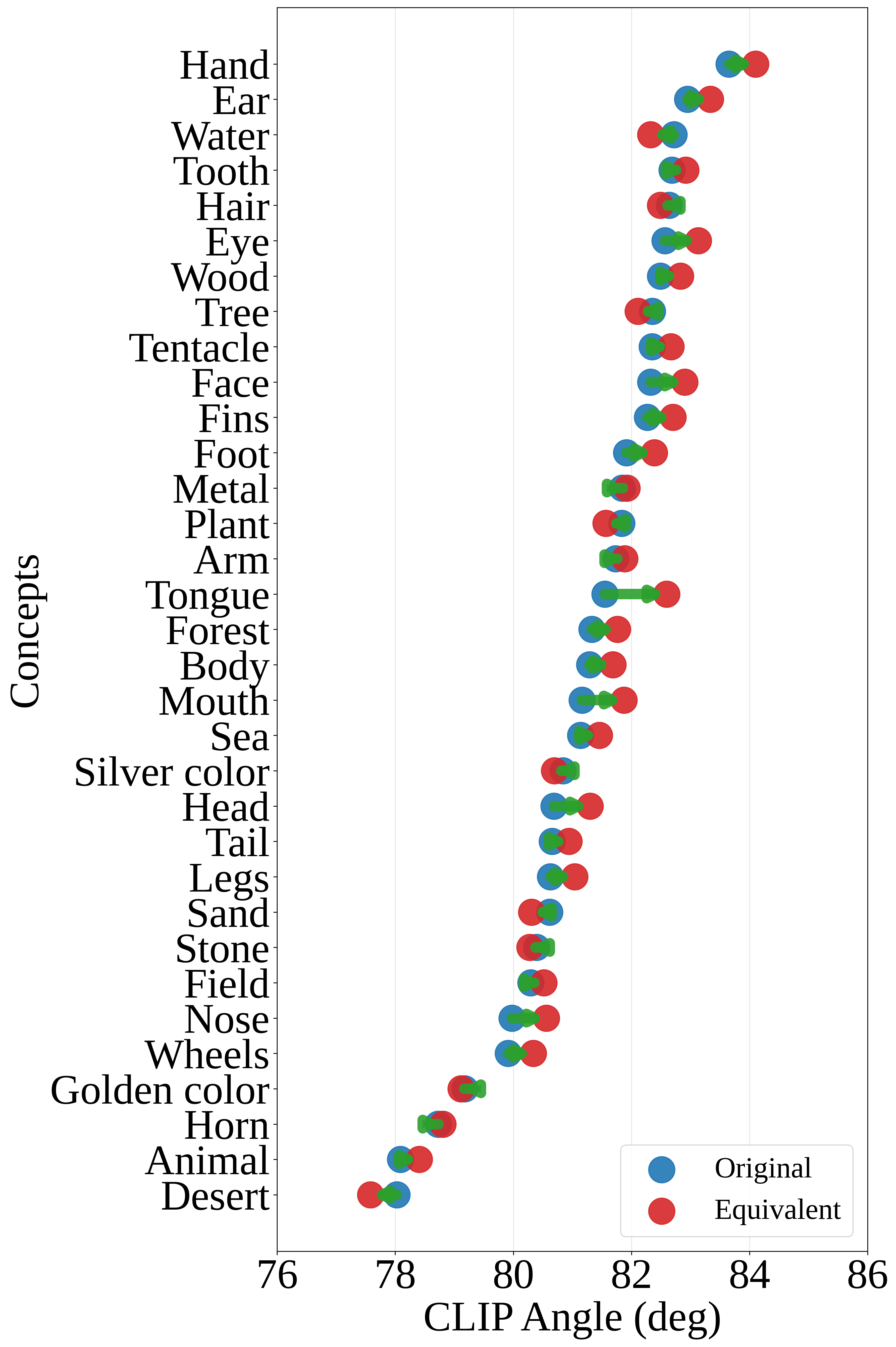}
  \caption{Open-vocabulary analysis for the ``Arabian Camel'' class in
  DinoViT.
  We show a larger set of prompts and their CLIP similarities to the class,
  illustrating the semantic neighborhood used in our analysis.}
  \label{fig:camel_full}
\end{figure}

\begin{figure}[]
  \centering
  \includegraphics[width=0.85\textwidth]{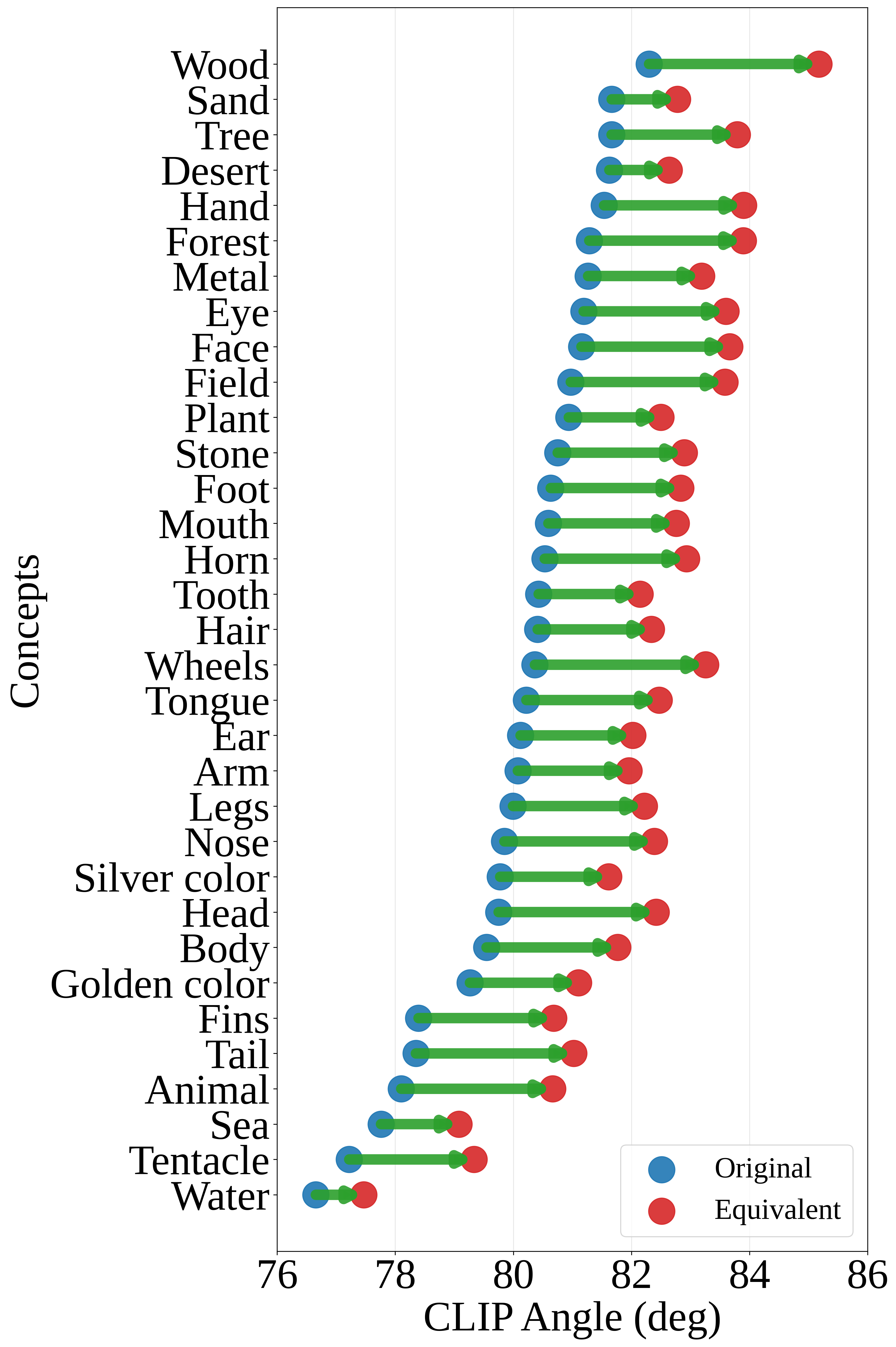}
  \caption{Open-vocabulary analysis for the ``Jellyfish'' class in DinoViT.
  The graph highlights related concepts and their CLIP similarities, showing
  that the concepts discussed in the main paper are part of a consistent
  semantic cluster.}
  \label{fig:jellyfish_full}
\end{figure}

We provide violin plots for all models that participated in our experiments
(\Cref{fig:bitresnet_violin,fig:resnext_violin,fig:efficientnet_violin}).
Each violin summarizes the distribution of semantic angle changes (in
degrees) under null-space perturbations for a given class.
\Cref{fig:camel_full,fig:jellyfish_full} provide extended open-vocabulary
concept lists used in the class analyses of the ``Arabian Camel'' and
``Jellyfish'' classes in DinoViT.
Nodes correspond to text prompts and the target class, and edge strengths
reflect CLIP similarity between image and text embeddings.
These plots show that the concepts we highlight in the main paper are
representative of broader open-vocabulary neighborhoods.

\section{DinoViT feature wrapper}
\label{app:dino_wrapper}

We use a wrapper around a pre-trained DinoViT backbone
\citep{caron2021emerging} to expose the penultimate feature $f$ and the
classifier head weights $W$.
We extract the sequence of tokens from the layer immediately before the
classifier head (denoted \texttt{"encoder.ln"} in our implementation),
take the class token as $f$, and apply the original head to obtain logits
$\ell(f) = W f$.

\begin{minipage}{0.68\linewidth}
\begin{lstlisting}
class SelectClassToken(nn.Module):
    def __init__(self, f):
        super().__init__()
        self.f, self.B = f, 1
    def forward(self, x):
        # x: (B * num_tokens, f); reshape and select class token (index 0)
        return x.reshape(self.B, -1, self.f)[:, 0, :]
    def set_B(self, B=1):
        self.B = B

class DinoHookable(nn.Module):
    def __init__(self, base: nn.Module, extractor, feature_dim=1024):
        super().__init__()
        self.extractor = extractor
        self.fc = base.heads.head        # classifier head; weights are W^T
        self.penultimate = SelectClassToken(f=feature_dim)
    def forward(self, x: torch.Tensor) -> torch.Tensor:
        self.penultimate.set_B(x.size(0))
        # token sequence before the classifier head
        x = self.extractor.extract(x, "encoder.ln")
        # penultimate feature f (class token)
        x = self.penultimate(x)
        # logits; penultimate feature f is available for analysis
        return self.fc(x)
\end{lstlisting}
\end{minipage}

This wrapper allows us to reuse the original DinoViT classifier while
directly accessing the feature space in which we construct translators and
null-space perturbations.




\end{document}